\documentclass{article}

\usepackage{arxiv}

\usepackage[utf8]{inputenc} % allow utf-8 input
\usepackage[T1]{fontenc}    % use 8-bit T1 fonts
\usepackage{hyperref}       % hyperlinks
\usepackage{url}            % simple URL typesetting
\usepackage{booktabs}       % professional-quality tables
\usepackage{amsfonts}       % blackboard math symbols
\usepackage{nicefrac}       % compact symbols for 1/2, etc.
\usepackage{microtype}      % microtypography
\usepackage{lipsum}
\usepackage{graphicx}
\usepackage{amsmath}
\usepackage{float}
\usepackage{subcaption}

\graphicspath{ {./images/} }

\title{Mapping the Edge of Chaos: 

Fractal-Like Boundaries in The Trainability of \\Decoder-Only Transformer Models}

\author{
 Bahman Torkamandi \\
  \texttt{bn.torkamandi@gmail.com} 
}

\begin{document}
\maketitle
\begin{abstract}

In the realm of fractal geometry, intricate structures emerge from simple iterative processes that partition parameter spaces into regions of stability and instability. 
Likewise, training large language models involves iteratively applying update functions, such as Adam, where even slight hyperparameter adjustments can shift the training process from convergence to divergence.
Recent evidence from miniature neural networks suggests that the boundary separating these outcomes displays fractal characteristics \cite{sohl2024boundary}.
Building on these insights, this study extends them to medium-sized, decoder-only transformer architectures by employing a more consistent convergence measure and examining the learning rate hyperparameter landscape for attention and fully connected layers. The results show that the trainability frontier is not a simple threshold; rather, it forms a self-similar yet seemingly random structure at multiple scales, with statistically consistent and repeating patterns. Within this landscape, a region of stable convergence is surrounded by a complex chaotic border, illustrating the sensitive nature of the underlying training dynamics.

\end{abstract}

%%keywords can be removed
\keywords{Attention, Transformers, Large Language Models, Chaos, Fractals, Neural Network Trainability}

\section{Introduction}

Fractals are intricate and infinitely complex patterns that can emerge from simple, $repeating$ $processes$. They are characterized by self-similarity, meaning their structure appears similar at any level of magnification. Fractals can be found both in nature such as in the branching of trees, the formation of snowflakes, and coastlines and in mathematical constructs. Their unique ability to model irregular and fragmented shapes makes them invaluable in various fields, including computer graphics, biology, and physics \cite{barnsley2014fractals}.

A fractal can be described as a set that exhibits \textbf{self-similarity} at every scale and possesses a \textbf{non-integer} Hausdorff dimension, which quantifies its complexity. Fractal \( F \) typically has features that satisfies the following conditions:
\begin{enumerate}
    \item \textbf{Similarity Transformations:} There exists a number of similarity transformations (scalings, translations, rotations) that cover the fractal \( F \) with scaled-down copies of itself.
    \item \textbf{Fine Structures:} Fractal \( F \)  contains detailed structure at arbitrary scales.
    \item \textbf{Hausdorff Dimension:} Most fractal have a Hausdorff dimension \( D \) that exceeds their topological dimension. The Hausdorff dimension is defined using the Hausdorff measure, which extends the concept of measuring lengths, areas, and volumes to non-integer dimensions \cite{falconer2013fractal}.
\end{enumerate}

Many fractals are generated through iterative processes. Iterated Function Systems (IFS) utilize the repeated application of contraction mappings to construct self-similar structures, as exemplified by the Sierpinski triangle \cite{barnsley2014fractals}. 
Some fractals, however, do not display exact self-similarity but rather exhibit quasi-self-similar patterns.
Another class of fractals is produced by iterating complex functions; for instance, the Multibrot set is generated by iterating the complex function
$f(z,c) = z^d + c$,
where integer $d\ge2$, and determining whether the sequence diverges or remains bounded based on different values of the complex parameter $c$ starting with $z_0=0$. Likewise, Multijulia sets are defined by fixing the parameter $c$ and varying the initial values of $z$ in the function
$f(z,c) = z^d + c$.
The boundary in the complex plane, separating initial $z$ values whose associated sequences remain bounded from those that diverge, exhibits a fractal structure.

\begin{figure}[h!]
    \centering
    % subfigure (a)
    \begin{subfigure}[b]{0.20\textwidth}
        \centering
        \includegraphics[width=\linewidth]{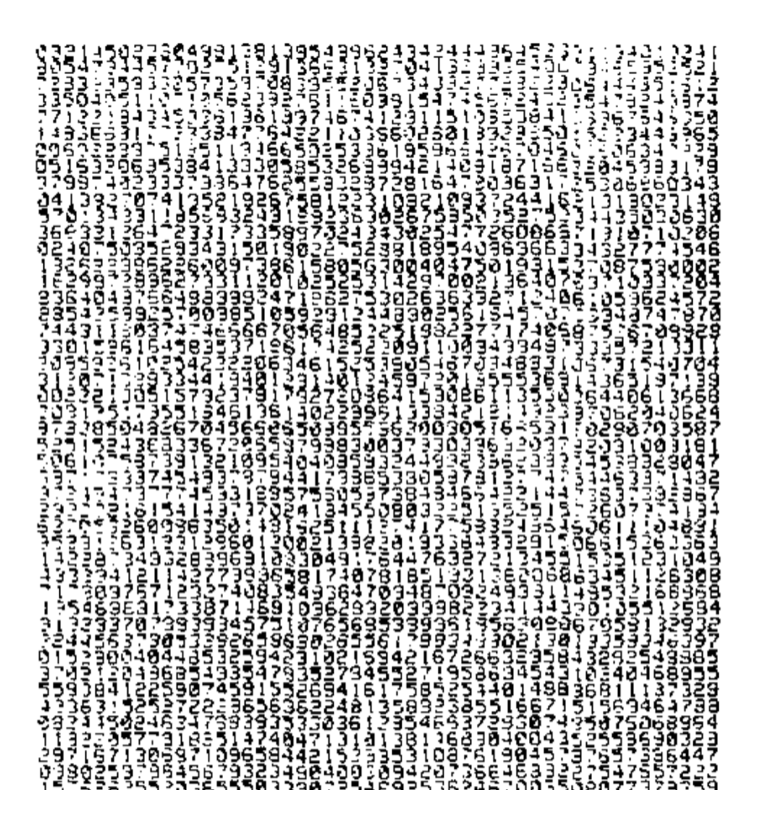}
        \caption{Random Number Table}
        \label{fig:rnt}
    \end{subfigure}
    \hfill
    % subfigure (b)
    \begin{subfigure}[b]{0.21\textwidth}
        \centering
        \includegraphics[width=\linewidth]{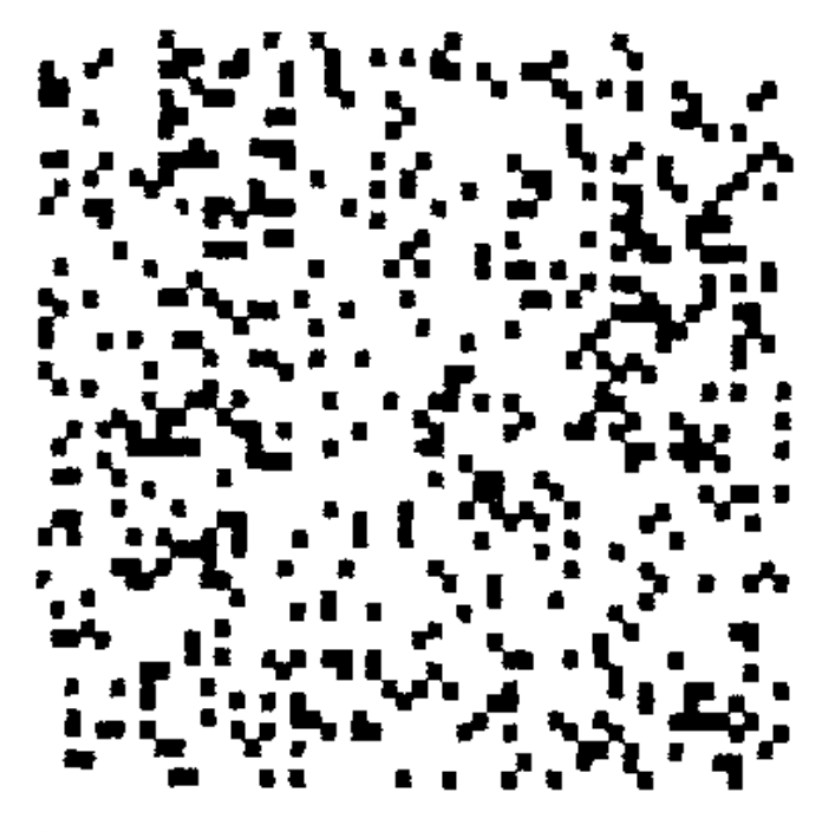}
        \caption{Simulated Coverage}
        \label{fig:dust}
    \end{subfigure}
    \hfill
    % subfigure (c)
    \begin{subfigure}[b]{0.16\textwidth}
        \centering
        \includegraphics[width=\linewidth]{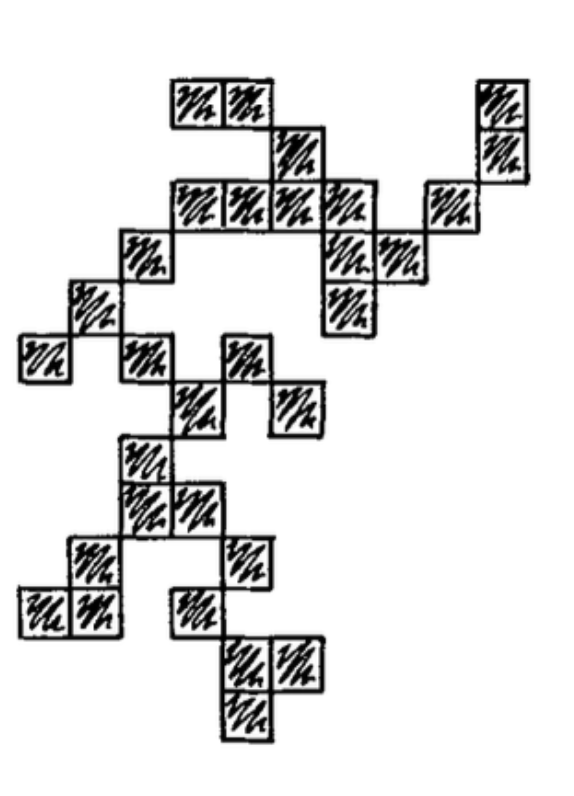}
        \caption{Largest Cluster}
        \label{fig:cluster}
    \end{subfigure}

    \caption{Fractal-like Dust Clusters From Monte Carlo Random Placements \cite{kaye1994random}.}
    \label{fig:three-panel}
\end{figure}

An intriguing real-world fractal example appears in chapter 5 of \cite{kaye1994random}, Fractal Systems Generated by Randomwalks in Two-Dimensional Space, fine particles arriving at a filter surface can be deposited on top of those collected earlier, producing clusters that significantly affect respirable dust assessments. These seemingly random deposits often exhibit pronounced clustering behavior. A straightforward way to visualize this effect is through a Monte Carlo simulation, in which entries from a random number table represent dust fineparticles, with both the random number table and resulting fractal structures shown in figure \ref{fig:three-panel}. Despite the underlying randomness, the resulting two-dimensional patterns can show fractal-like clusters reminiscent of random walks, as repeated random placements can yield clumping structures. Such overlaps raise the crucial question of whether a cluster on the filter consists of one large non-respirable agglomerate or multiple smaller respirable fineparticles; a misclassification can introduce considerable errors into estimations of respirable dust hazards \cite{kaye1994random}.

Newton fractals provide another striking example of fractal geometry emerging from iterative processes. These fractals arise from applying Newton’s method to find roots of complex functions. Iterating from different initial points partitions the complex plane into basins of attraction, each converging to a different root. The intricate, fractal boundaries between these basins highlight the sensitivity and complexity inherent in such iterative processes \cite{peitgen2004chaos}.

Training neural networks involves $iterative$ $updates$ of model parameters\textemdash commonly through optimization algorithms like gradient descent or Adam\textemdash which can lead to either convergent or divergent outcomes. This process is highly sensitive to hyperparameters, and small adjustments can significantly impact the stability of training.

When training a neural network, we iterate a function (e.g., a gradient descent step) involving many variables (the parameters of the neural network). For instance, if we perform full-batch gradient descent with a fixed learning rate $\eta$  , we update the parameters 
W by iterating the function:

$$W_{t+1}=W_{t}- \eta \nabla L(W_{t})$$

where $ \nabla L(W_t)$ 
is the gradient of the training loss at iteration $t$. 
More advanced optimization algorithms like Adam \cite{kingma2014adam} incorporate adaptive learning rates and momentum terms. The Adam optimizer updates the parameters using estimates of the first and second moments of the gradients:

\begin{align*} 
    m_{i}^{t+1} &= \beta_1 m_i^t + (1 - \beta_1) \frac{\partial L(W_t)}{\partial w_i}  \\
    v_i^{t+1} &= \beta_2 v_i^t + (1 - \beta_2) (\frac{\partial L(W_t)}{\partial w_i})^2 \\
    \hat{m}_i^{t+1} &= \frac{m_i^{t+1}}{1 - \beta_1^{(t+1)}} \\
    \hat{v}_i^{t+1} &= \frac{v_i^{t+1}}{1 - \beta_2^{(t+1)}} \\
    w_i^{t+1} &= w_i^t - \eta (\frac{\hat{m}_i^{t+1}}{\sqrt{\hat{v}_i^{t+1}} + \epsilon}) 
\end{align*}

where 
$\hat{m}_i^t$	
  and 
$\hat{v}_i^t$
  are the estimates of the first and second moments, 
$\beta_1$ and $\beta_2$ are exponential decay rates, and 
$\epsilon$ is a small constant for numerical stability.

Recently, transformer models have revolutionized natural language processing (NLP) by enabling efficient handling of sequential data through self-attention mechanisms. As these models scale to billions of parameters, training them effectively becomes increasingly challenging. Issues such as vanishing or exploding gradients, sensitivity to hyperparameters, and non-convex optimization landscapes complicate the training process.
Recent work by Sohl-Dickstein \cite{sohl2024boundary} demonstrated that the boundary between hyperparameters leading to successful and divergent training in very small neural networks exhibits fractal characteristics. This implies that small changes in hyperparameters can lead to significant differences in training outcomes, highlighting the complex and sensitive nature of the trainability landscape. Understanding this boundary is crucial for developing robust training methodologies.

In this paper, inspired by the iterative processes used to generate fractals, I examine the fractal characteristics of the hyperparameter boundaries that separate stable from divergent training regimes in large decoder-only transformer models optimized by algorithms such as Adam. Drawing on the parallels between fractal generation and neural network training, I present empirical evidence of repetitive structures embedded within the hyperparameter landscape.

%\clearpage

\section{Related work}
\label{sec:headings}
\subsection{Fractal Boundaries in Neural Networks}

Sohl-Dickstein \cite{sohl2024boundary} demonstrated that the boundary between hyperparameters leading to successful and divergent training in shallow neural networks exhibits fractal characteristics. By visualizing training outcomes over grids of hyperparameters, the study revealed that the boundary is intricate and self-similar across different scales and training conditions \cite{sohl2024boundary}. 

In his experiments, the author trained a single hidden layer neural network comprising 16 neurons using gradient descent and MSE loss, utilizing activation functions such as \texttt{tanh} or \texttt{ReLU}. The training dataset contained the same number of examples as the number of free parameters in the network, totaling 272. The inputs were initialized from a standard normal distribution.

To assess convergence, the loss was scaled to start at 1. Training was considered to have converged if the mean of the last 20 loss values fell below 1 \cite{sohl-dickstein_fractal}. The convergence measure was defined as follows:

\[
\text{convergence\_measure} =
\begin{cases}
\displaystyle \sum_{i} l_i, & \text{if converged}, \\
\displaystyle \sum_{i} \frac{1}{l_i}, & \text{otherwise}.
\end{cases}
\]

He reported fractal dimensions ranging from \(1.17\) to \(1.98\) across various configurations. Specifically, a fractal dimension of \(1.55\) was observed when using a minibatch with the \texttt{tanh} activation, and learning rates for both the input and hidden layers as hyperprameters.

\subsection{Transformer Models} 

Introduced by Vaswani et al. \cite{vaswani2017attention}, transformer architectures have reshaped language models by enabling parallel processing of sequential data through self-attention mechanisms. This paradigm shift allows models to capture long-range dependencies more effectively than traditional recurrent architectures. Despite their success, training large transformers remains challenging due to their scale and the complexity of their optimization landscapes.
Building on the foundational transformer architecture, various adaptations and enhancements have been proposed to address specific limitations and extend their capabilities. Notably, decoder-only transformer models, such as the Generative Pre-trained Transformer (GPT) series \cite{radford2018improving,radford2019language, brown2020language}, have gained significant attention for their prowess in generative tasks.

Although transformers are most prominently used for text, they have been extended to other domains, including computer vision, speech recognition, and multimodal tasks. In computer vision, Vision Transformers (ViT) have achieved competitive results by treating image patches as “tokens” and applying attention across them, demonstrating performance on par with or exceeding traditional convolutional networks \cite{alexey2021image}. Similarly, transformer-based approaches in speech recognition, such as Speech-Transformer, leverage the self-attention mechanism to model acoustic sequences effectively \cite{dong2018speech}.

Despite their architectural advantages, scaling transformer models to handle vast datasets and complex tasks introduces significant challenges. Large-scale transformers, encompassing billions of parameters, demand substantial computational resources and memory bandwidth, making their training both time-consuming and economically demanding. Additionally, the optimization landscape of such expansive models is fraught with difficulties, including numerous local minima and saddle points that can hinder effective convergence.

To address these challenges, several strategies have been employed:

\begin{itemize}
    \item {Distributed Training and Model Parallelism}: Techniques such as tensor parallelism, pipeline parallelism, and data parallelism enable the distribution of model training across multiple GPUs or even clusters of machines, thereby alleviating memory constraints and reducing training times.
    
    \item {Advanced Optimization Algorithms}: Beyond standard optimization methods like Adam, newer algorithms such as LAMB (Layer-wise Adaptive Moments optimizer for Batch training) \cite{you2019large} have been proposed to enhance convergence rates and stability during the training of large-scale transformers.
    
    \item {Regularization Techniques}: Incorporating regularization methods like dropout and weight decay helps prevent overfitting and improves the generalization capabilities of transformer models.
\end{itemize}

\section{Network Architecture}

In this study, I employ a decoder-only transformer architecture tailored for autoregressive character prediction. Unlike the standard transformer model, which consists of both encoder and decoder stacks, the model utilizes exclusively decoder layers to generate the next character in a sequence based on the preceding context. Each decoder layer incorporates multi-head self-attention mechanisms and feedforward networks, complemented by layer normalization and residual connections to ensure stable and efficient training. The model comprises $95,973$ trainable parameters.

The network configuration is as follows:
\begin{itemize}
    \item \textbf{Context Length}: 64 tokens/characters.
    \item \textbf{Temperature}: 0.3.
    \item \textbf{Mini Batch Size}: 256.
    \item \textbf{Attention Key-Query Dimension}: 64 (matching the model embedding dimension).
    \item \textbf{Positional Encoding}: Utilizes a non-trainable sinusoidal positional encoder to preserve the sequential information of input characters, enabling the transformer to process data in an autoregressive manner.
    \item \textbf{Initialization}:
    The weights in the neural network are initialized using default initializers provided by Flax library, with the same initialization used each time the network is trained.
    \item \textbf{ Loss function }:
     Loss function is the softmax cross-entropy loss.
\end{itemize}

\clearpage

\begin{figure}[htbp]
  \centering
  \includegraphics[width=14cm]{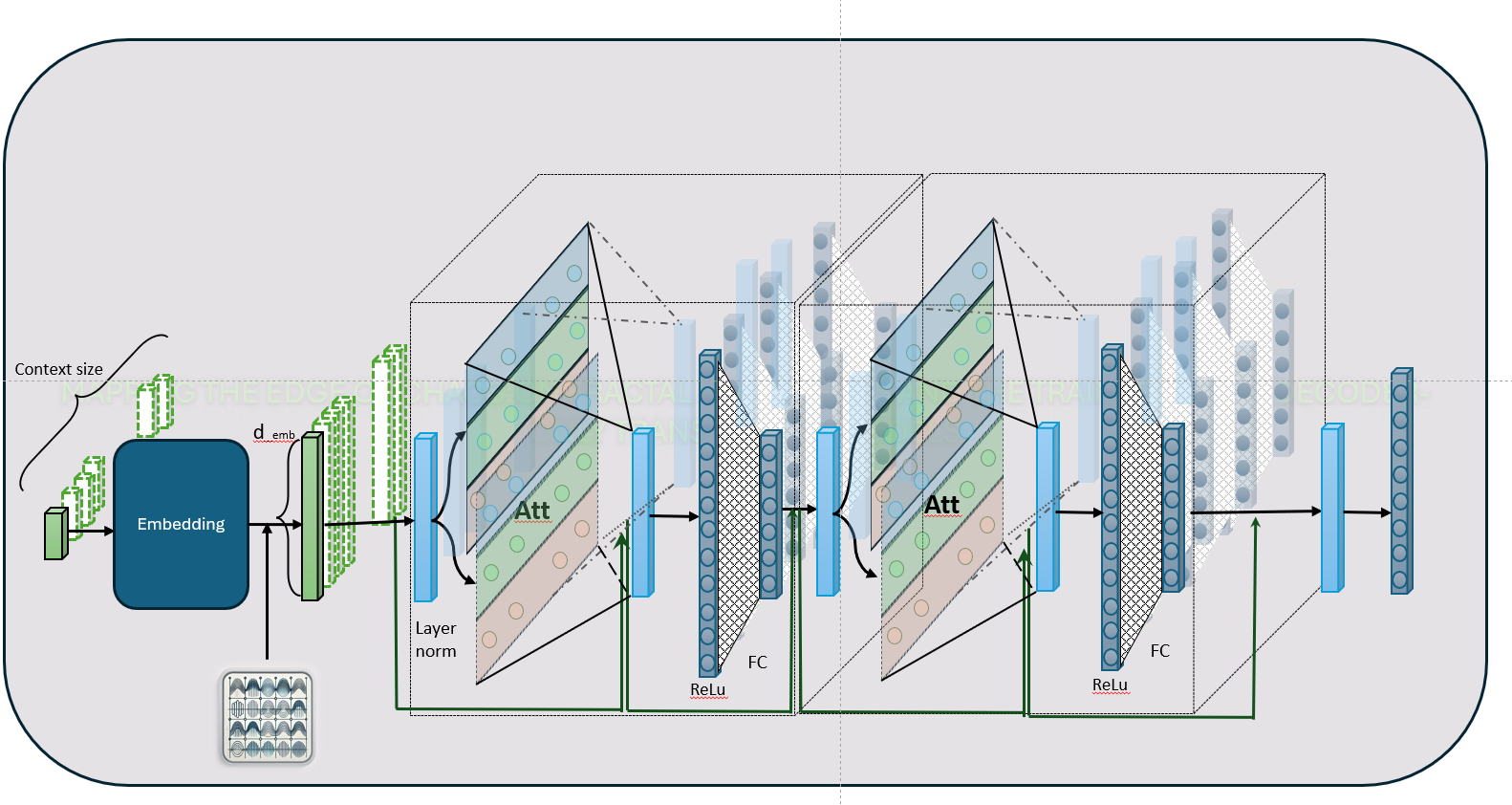} 
  \caption{Architecture of The Language Model.}
  \label{fig:topology}
\end{figure}

\noindent The detailed network architecture is outlined below (Figure \ref{fig:topology}):
\begin{itemize}
    \item \textbf{Input Tokens}: Sequences of token with a length of 64.
    \item \textbf{Embedding Layer}:
    \begin{itemize}
        \item Maps tokens to embeddings.
    \end{itemize}
    \item \textbf{Positional Encoding}:
    \begin{itemize}
        \item Adds positional information to embeddings.
    \end{itemize}
    \item \textbf{Transformer Layers (2 layers)}:
    \begin{itemize}
        \item \textbf{Within Each Layer}:
        \begin{itemize}
            \item \textbf{\textit{Layer Normalization}}
            \item \textbf{\textit{Two-Head Self-Attention}}
            \item \textbf{Residual Connection}
            \item \textbf{Layer Normalization}
            \item \textbf{Feed-Forward Network} (2 layers with ReLU and linear activations).
            \item \textbf{Residual Connection}
        \end{itemize}
    \end{itemize}
    \item \textbf{Final Layer Normalization}:
    \begin{itemize}
        \item Normalizes the output from the last transformer layer.
    \end{itemize}
    \item \textbf{Output Layer}:
    \begin{itemize}
        \item Projects normalized outputs to vocabulary logits.
    \end{itemize}
\end{itemize}

%\clearpage

\noindent The model is trained using Adam, fixed initialization and data, softmax cross-entropy loss, a context size of 64 tokens/characters and a batch size of 256. The key-query dimension in the attention layers is set to 64, aligning with the model's embedding dimension. A non-trainable sinusoidal positional encoder is integrated to maintain the sequential order of input characters, facilitating autoregressive data processing. The embedding is learned during training. Attention layers and their layer norm are trained with learning rate $\eta_{att}$ and the rest of the trainable parameters are trained using learning rate $\eta_{fc}$.

\subsection{Data}

The dataset comprises $1,071,890$ strings, each containing 64 characters, extracted from the Complete Works of William Shakespeare available on Project Gutenberg \cite{shakespeare_complete_works}. I use characters as tokens, resulting in a vocabulary size of $101$ unique tokens.

\clearpage

\section{Convergence Measure}

The convergence criterion and the convergence measure in Sohl-Dickstein \cite{sohl2024boundary} work is not sufficient nor consistent, therefore the model is considered to have converged when:
\begin{itemize}
    \item The mean of recent losses is below 0.4 (determined based on the data and generation quality).
    \item The mean of recent losses is at least 0.1 lower than the mean of early losses.
    \item The variance of recent losses is below a threshold of 0.01.
\end{itemize}
I use the first and last $5\%$ of the training steps to calculate early and late losses, respectively.

To investigate the trainability boundary, I evaluate whether training runs converge or diverge across various hyperparameter ranges, focusing on the learning rates for fully connected layers and attention mechanisms. First, I establish a more concrete definition of convergence based on the loss function derived from hypothetical training scenarios that represent the most convergent and divergent behaviors. I then convert the convergence metric to a color map, which visually represents the degree of convergence. This approach allows us to identify abrupt state changes and discern any chaotic or transitional phases in trainability. By mapping these variations, I can effectively visualize and analyze the stability of the training process under different learning rate settings.

\begin{figure}[htbp]
  \centering
  \includegraphics[width=12cm]{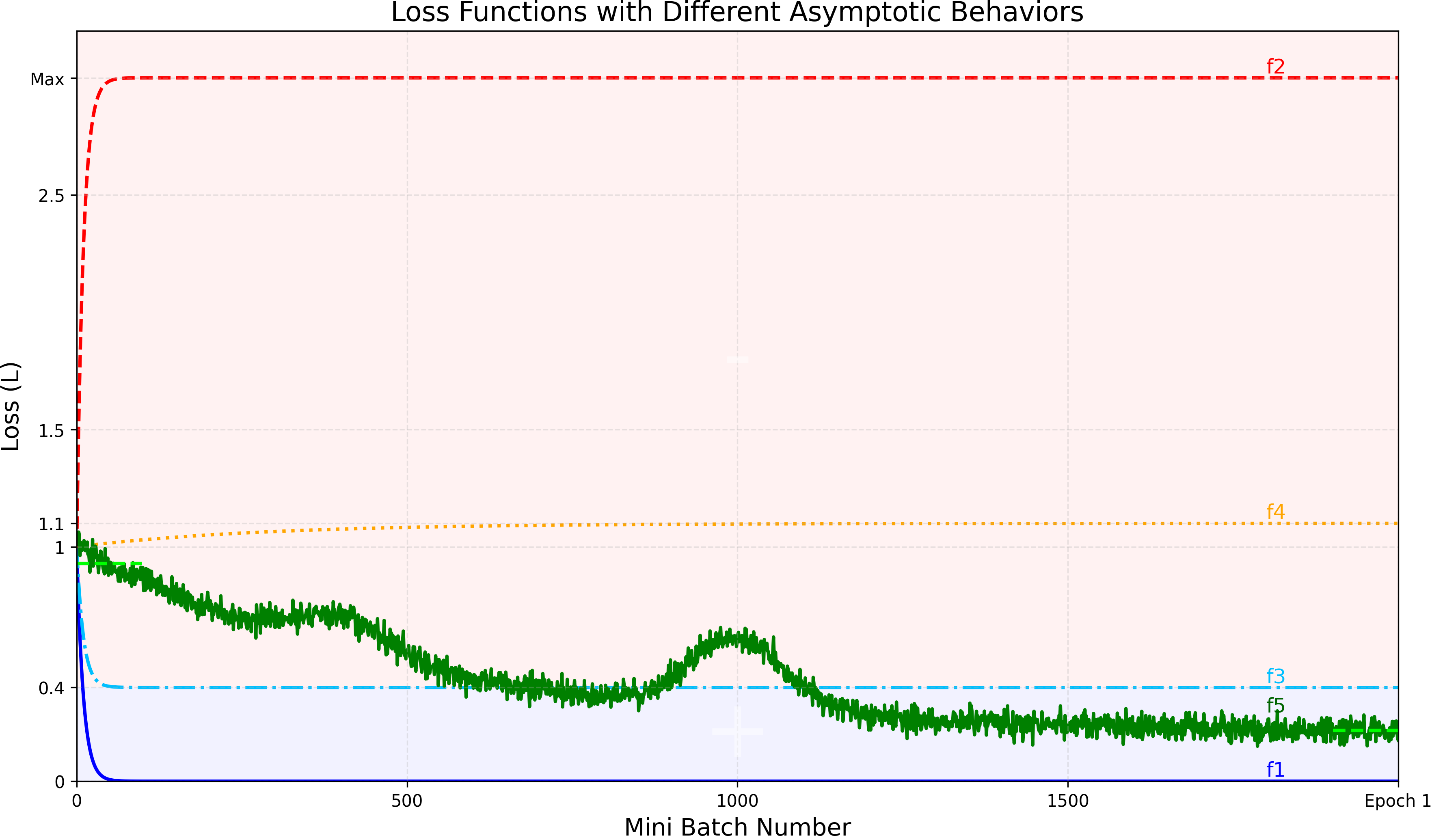} 
  \caption{Loss Functions Illustrating Various Convergence Behaviors.}
  \label{fig:fig2}
\end{figure}

Figure \ref{fig:fig2} depicts five hypothetical loss functions $f_1$, \dots, $f_5$. The losses are normalized with respect to the loss at step 1 (minibatch 1). $f_1$ is the most convergent and $f_2$ is the most divergent, whereas $f_4$ exhibits moderate divergence. I assume that for any given network and training data, a supremum exists and is denoted by $Max$. Additionally, the graph 
$f_3$ is identified as both the least convergent and the least divergent graph.

The degree of convergence should be proportional to the area (or the sum of losses for practical purposes) between the loss curve and the least convergence/divergence graph, $f_3$. I establish the convergence measure to be 1 for $f_1$ and -1 for $f_2$, while the least convergent/divergent loss function has a convergence measure of 0. Formally, I define the convergence measure as follows:

Let:
\begin{flalign*}
    N &= \text{len(normalized\_losses)} &&\\
    C &= \text{cut off} = 0.4 \\
    M &= \text{Max} \\
\end{flalign*}
then $1 + (N - 1)C$ is  the discrete approximation to the area under $f_3 $
. The convergence measure \( \mu \) is defined as:
\[
\mu =
\begin{cases}
\sqrt{ \dfrac{1 + (N - 1)C - \sum_{i=1}^{N} l_i }{(N - 1)C} }, & \text{if } \text{converged and} \sum_{i=1}^{N} l_i \leq 1+(N - 1) C , \\
- \sqrt{ \dfrac{ \sum_{i=1}^{N} l_i -(1 + (N - 1)C) }{1 + (N - 1)(M - C)} }, & \text{otherwise}.
\end{cases}
\]

where $l_i$ is each element in normalized losses.

This formulation constrains the convergence measure to be between -1 and 1, where -1 represents the most divergent and 1 represents the most convergent. This consistent measure allows us to use a color map proportional to the convergence measure to examine the trainability. I assign the most intense blue to 1 and the most intense red to -1, with  white representing 0.

I also calibrated this convergence measure using the output of the language model. Due to predicting next character, limited computational resources, fewer parameters, and less training data, developing a fully-fledged large language model (LLM) for English was not the focus. However, the model’s ability to generate English-like words instead of random or repetitive characters indicates that it has achieved meaningful convergence and has learned from the data.

Examples of generated response and the corresponding convergence measures for the language model tasked to complete "To be or not to be":

\begin{itemize}
    %%%
    \item $\mu$: -0.2494 \\
    To be or not to be oq;ëpy;;E*E;ëEpppAW1ëëEqXëWpE((ëqqEq(Wœp*Wqqëqqqœqë((ë:ëëqqqqWq1
    \item $\mu$: -0.1313 \\
    To be or not to be ther the the the and that the and the the the athe at I a h
     \item $\mu$: 0.1714 \\
    To be or not to be the painter the the are fartions the the so the ounlllllllllllll
    \item $\mu$: 0.4769 \\
    To be or not to be the partician of the soul and may and so some and thainomalovima 
    \item $\mu$: 0.4883\\
     To be or not to be the partice. \\
     \\
        CHIEF JUSTICE. \\
        I will be so your court trastristri
\end{itemize}

\section{Results}

Using the more rigorous convergence criterion and a consistent convergence measure, I trained the language model (from scratch) across a range of learning rates, varying both the attention layers and all other layers. This analysis revealed self-similar patterns at multiple scales, with repeating statistical characteristics at different levels of magnification.

\begin{figure}[h] % Use [H] to force the exact placement
  \centering
  \begin{minipage}{0.48\textwidth}
    \centering
    \includegraphics[width=\linewidth]{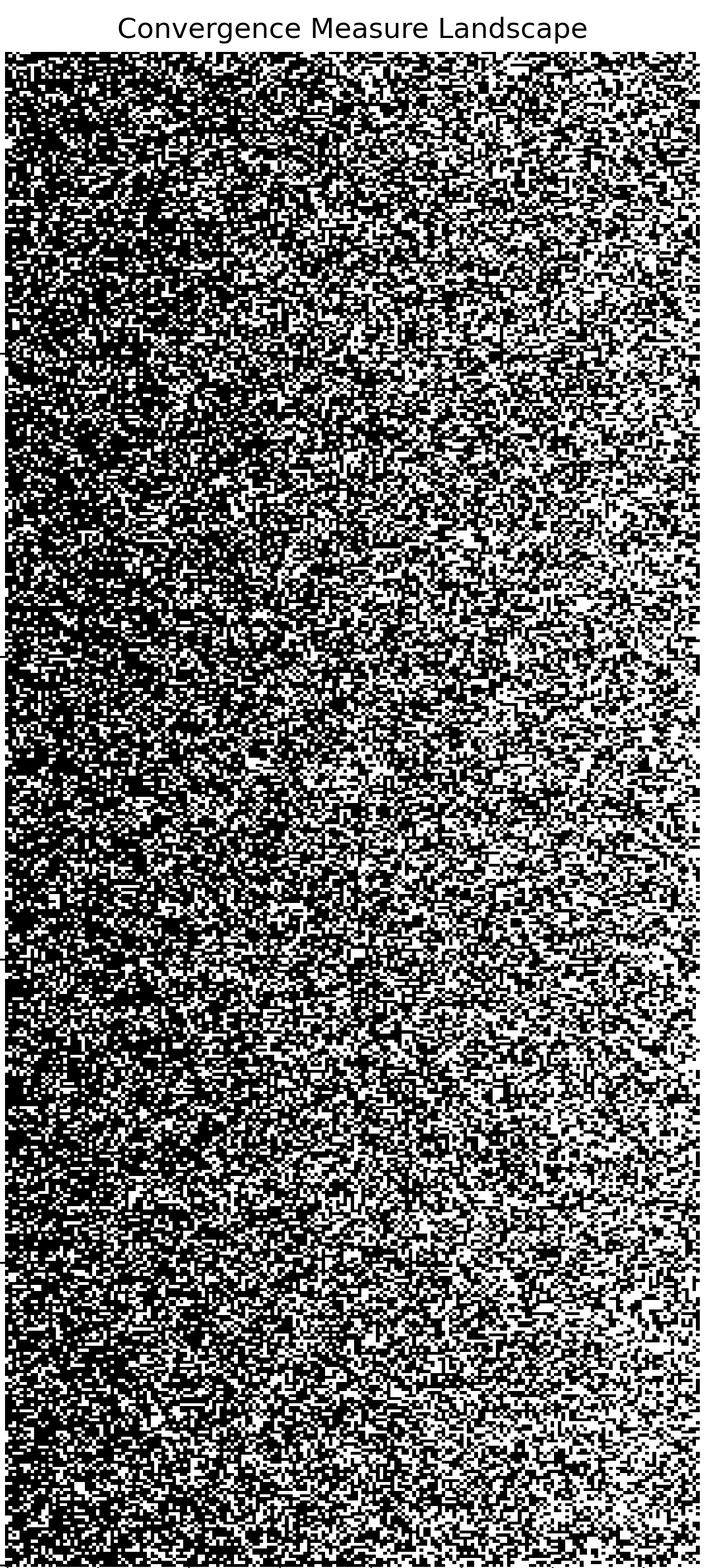}
    \caption{Convergence Measure Binary Heatmap, Granularity $10^{-5}$.}
    \label{fig:real_heat}
  \end{minipage}
  \hfill
  \begin{minipage}{0.51\textwidth}
    \centering
    \includegraphics[width=\linewidth]{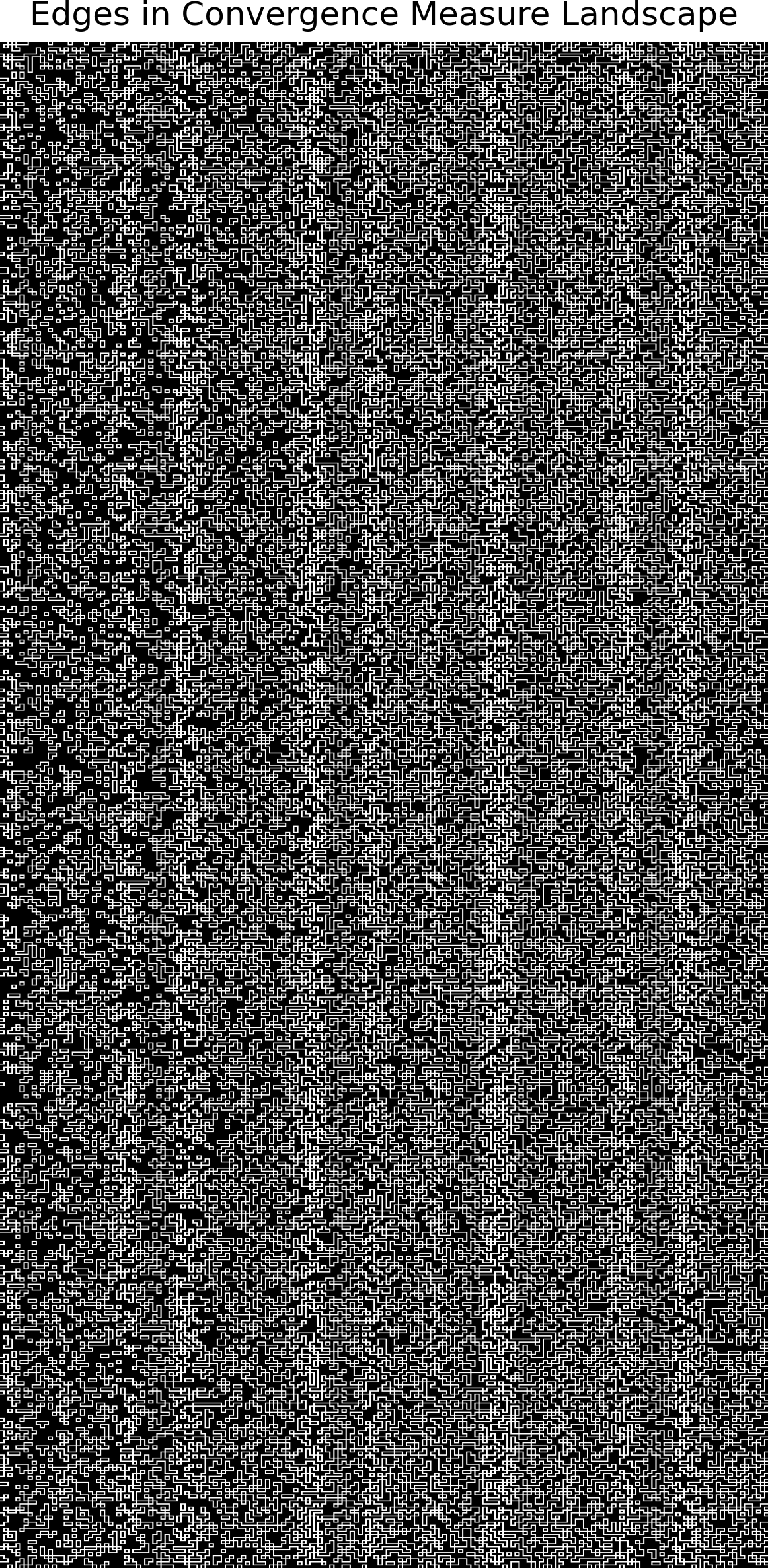}
    \caption{Boundaries Between Convergence and Divergence Regions, Granularity $10^{-5}$, Box-Count Dimension: $1.9772$.}
    \label{fig:real_edge}
  \end{minipage}
\end{figure}

In figure~\ref{fig:real_heat}, the regions that converged are shown in black and figure~\ref{fig:real_edge} presents the corresponding borders; these intricate boundaries have a dimension of $1.9772$. The intricate patterns observed throughout the hyperparameter convergence landscape underscore the complexity of trainability dynamics at a granularity of $10^{-5}$.

\begin{figure}[H]
  \centering
  \includegraphics[width=7.5cm]{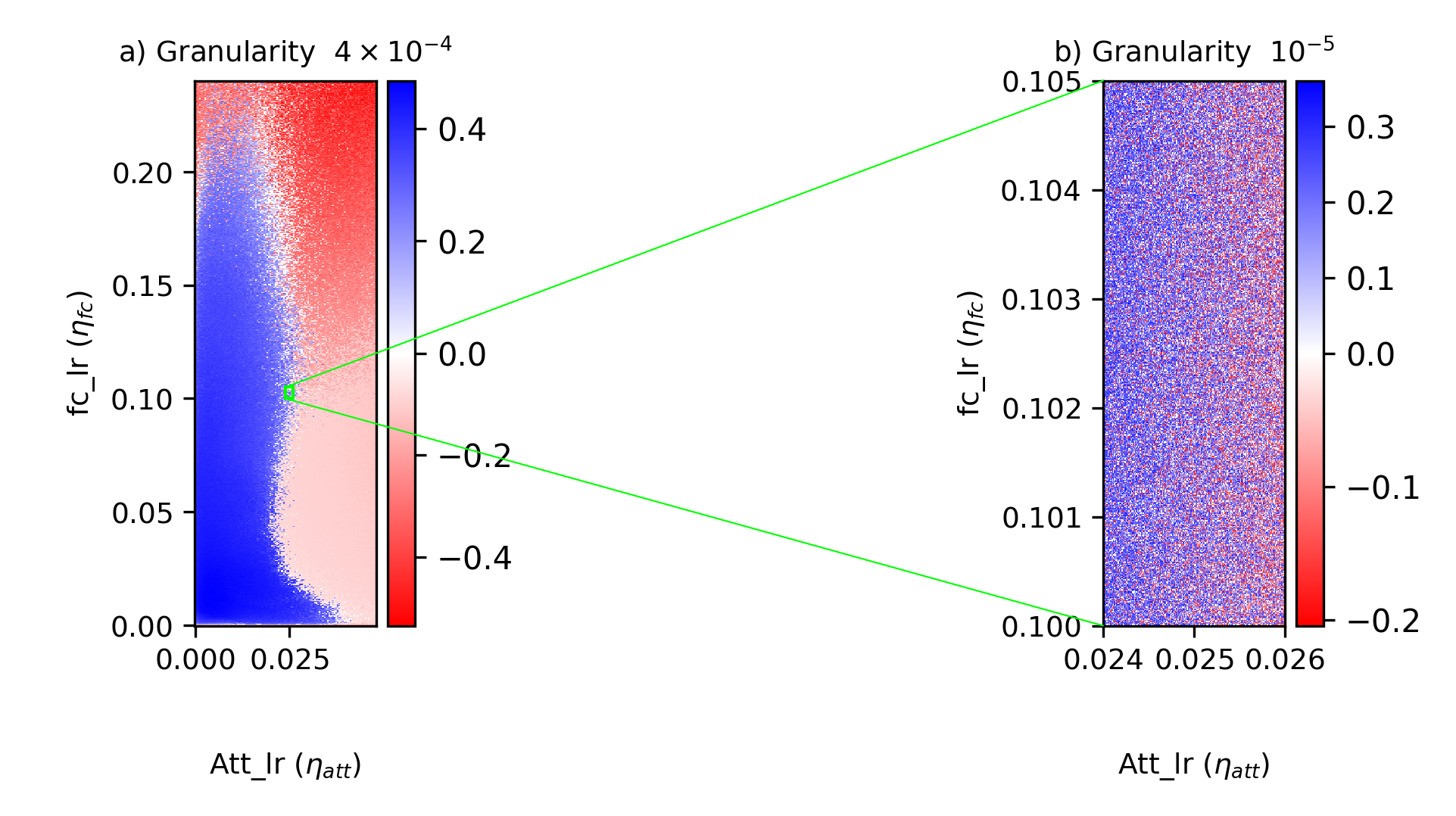} 
  \caption{Convergence Measure Heatmap – Boundaries at Granularity $10^{-5}$ Have a Dimension of $1.9772$.}
  \label{fig:10e-5}
\end{figure}

\begin{figure}[H]
  \centering
  \includegraphics[width=7.5cm]{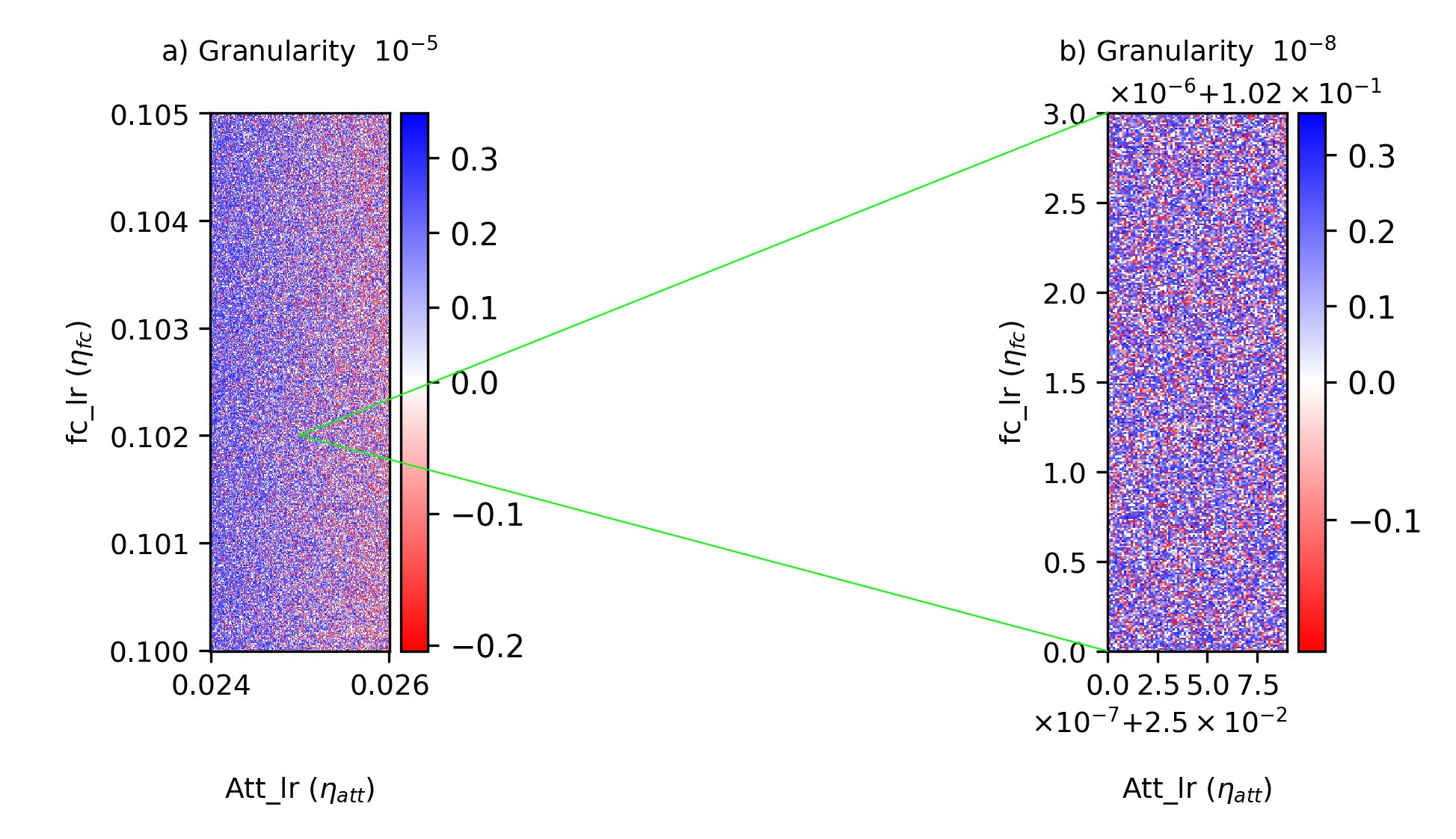} 
  \caption{Convergence Measure Heatmap – Boundaries at Granularity $10^{-8}$ Have a Dimension of $1.9715$.}
  \label{fig:10e-8}
\end{figure}

\begin{figure}[H]
  \centering
  \includegraphics[width=7.5cm]{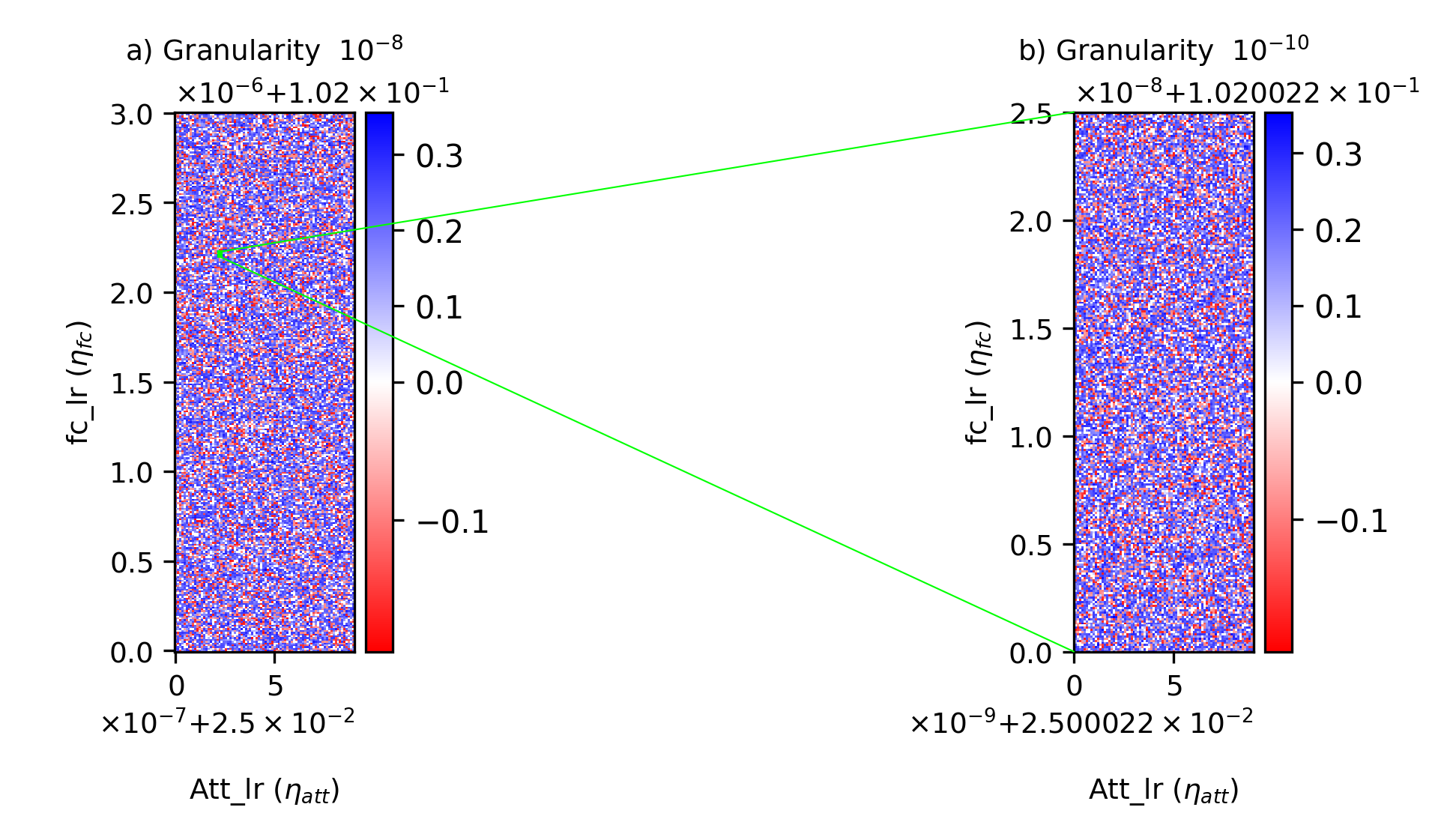} 
  \caption{Convergence Measure Heatmap – Boundaries at Granularity $10^{-10}$ Have a Dimension of $1.9649$.}
  \label{fig:10e-10}
\end{figure}

\begin{figure}[H]
  \centering
  \includegraphics[width=7.5cm]{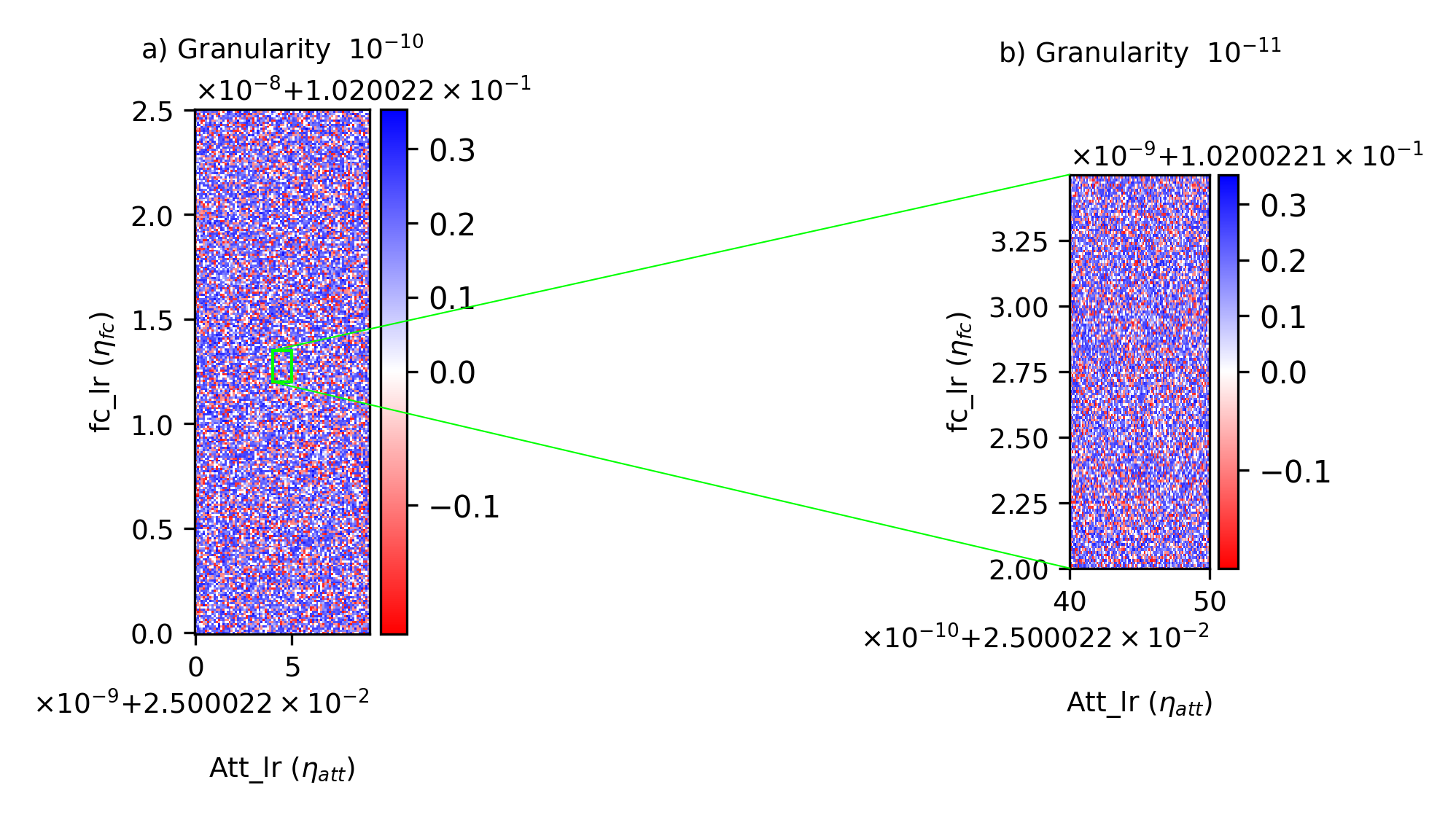} 
  \caption{Convergence Measure Heatmap – Boundaries at Granularity $10^{-11}$ Have a Dimension of $1.9783$.}
  \label{fig:10e-11}
\end{figure}

Figure~\ref{fig:10e-5}-a shows the hyperparameter convergence landscape for the learning rates of the attention layers and their layer normalization ($\mu_{att}$) versus all other layers ($\mu_{fc}$). In this figure, a stable region is surrounded by chaotic patterns. Zooming into a specific region (Figure~\ref{fig:10e-5}-b and ~\ref{fig:real_heat}) reveals intricate divergent and convergent behaviors, indicating a chaotic boundary. Boundaries in this zoomed region have a fractal dimension of $1.9772$. Further magnification (Figure~\ref{fig:10e-8}-b) uncovers similar patterns; this area, with a fractal dimension of $1.9715$, shows statistical self-similarity across scales and chaotic tendencies. Figure~\ref{fig:10e-10} likewise, demonstrates self-similar characteristics\textemdash as evidenced by similar textures and convergence-measure histograms at a Granularity of $10^{-10}$\textemdash which also has edges with a fractal dimension of $1.9649$. Similarly, figure~\ref{fig:10e-11} reproduces these textures and statistical properties, the edges show a box-count dimension of $1.9783$ (Figure~\ref{fig:10e-11}-b).

\begin{figure}[H]
  \centering
  \includegraphics[width=5.5cm]{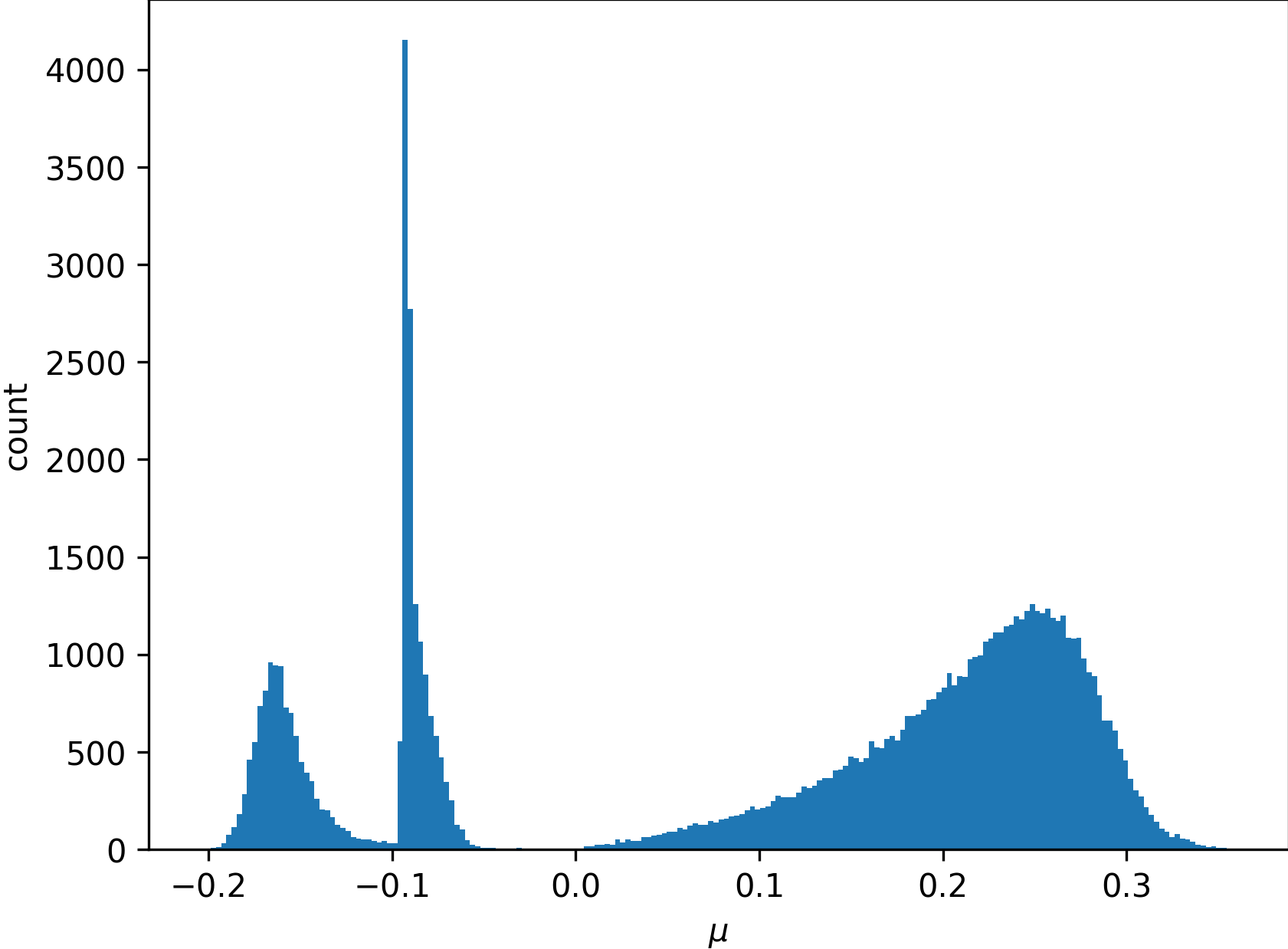} 
  \caption{Histogram of Convergence Measure, Granularity $10^{-5}$.}
  \label{fig:hist10m5}
\end{figure}

\begin{figure}[H]
  \centering
  \includegraphics[width=5.5cm]{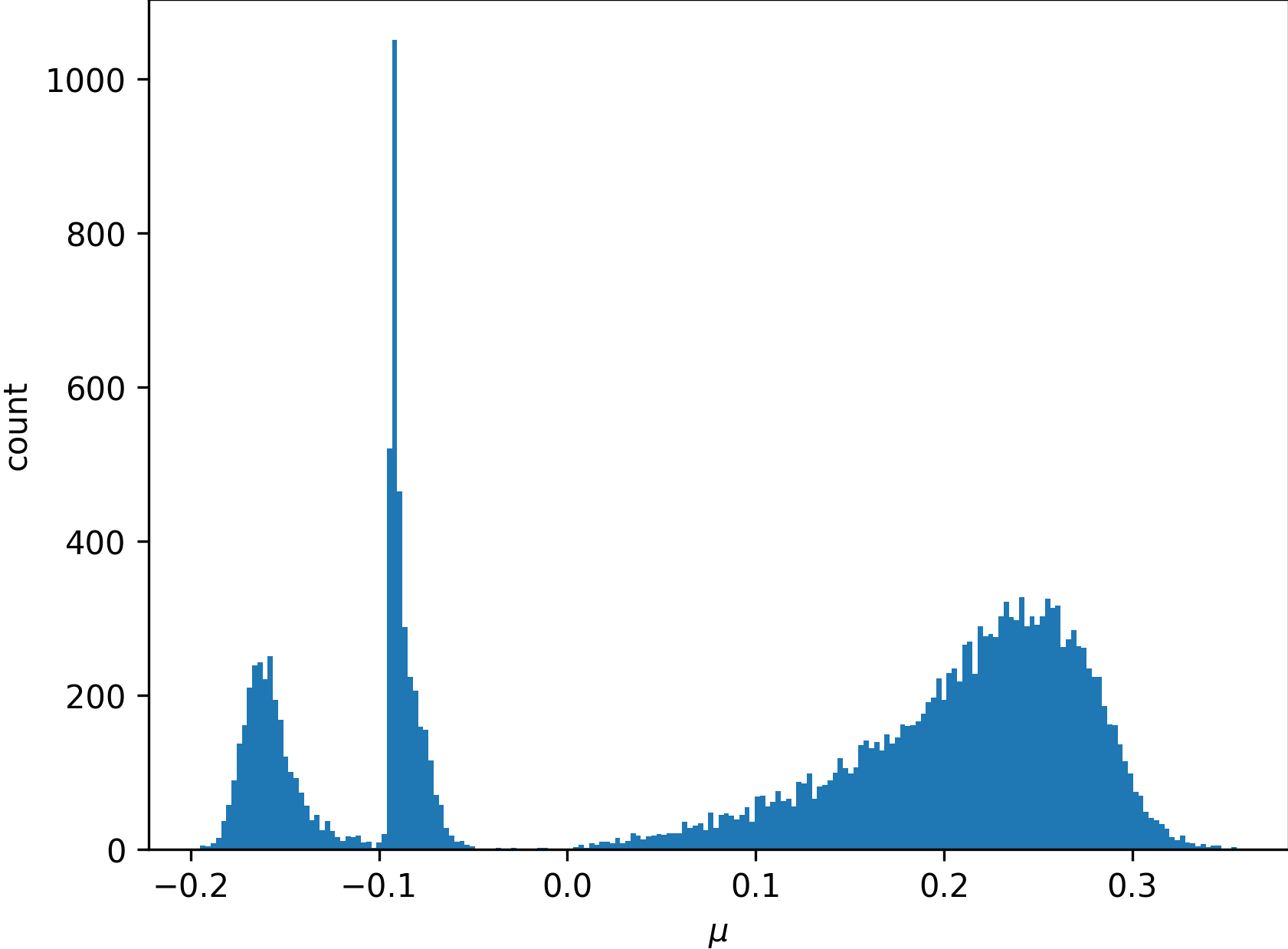} 
  \caption{Histogram of Convergence Measure, Granularity $10^{-8}$.}
  \label{fig:hist10m8}
\end{figure}

\begin{figure}[H]
  \centering
  \includegraphics[width=5.5cm]{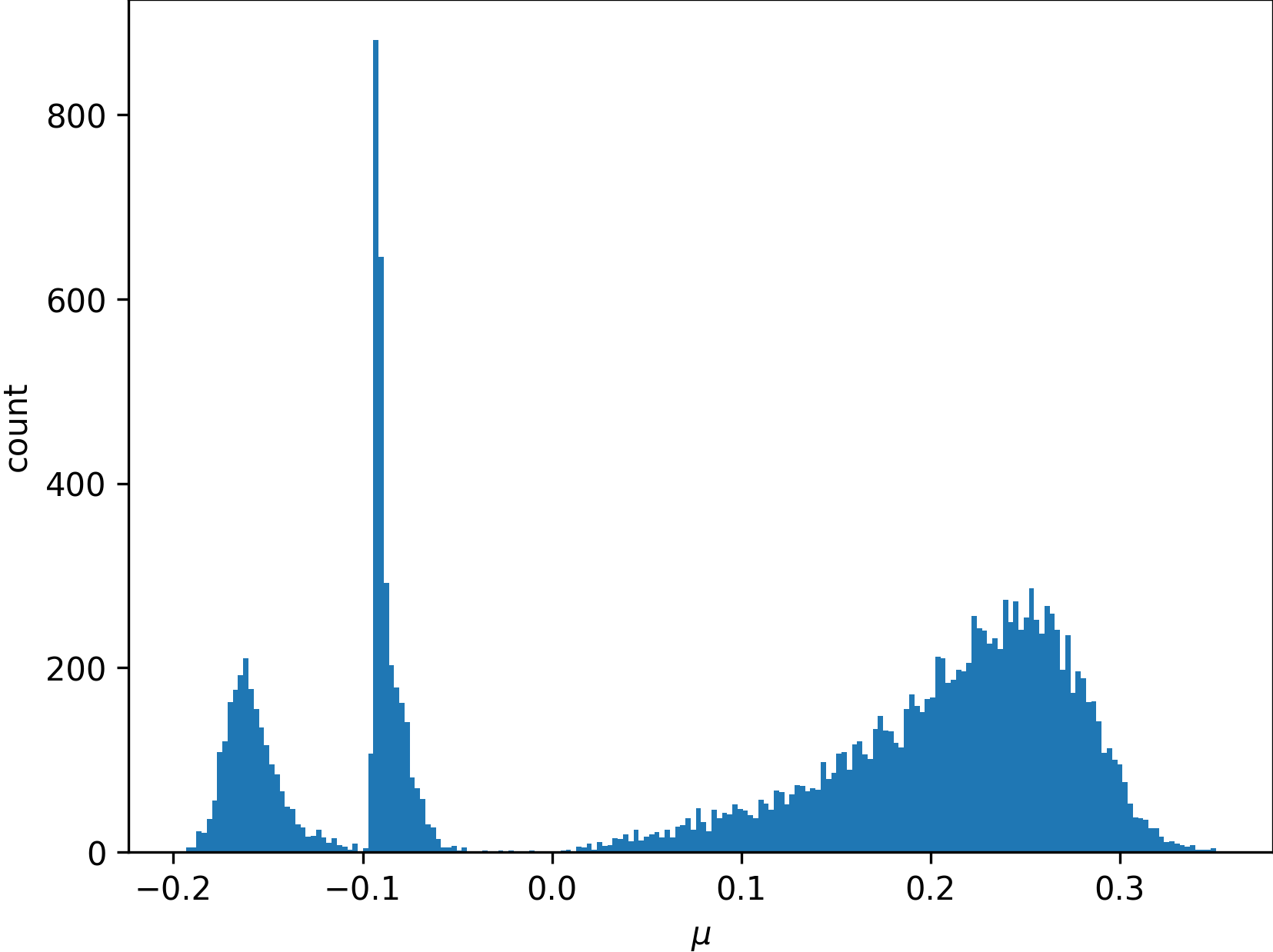} 
  \caption{Histogram of Convergence Measure, Granularity $10^{-10}$.}
  \label{fig:hist10m10}
\end{figure}

\begin{figure}[H]
  \centering
  \includegraphics[width=5.5cm]{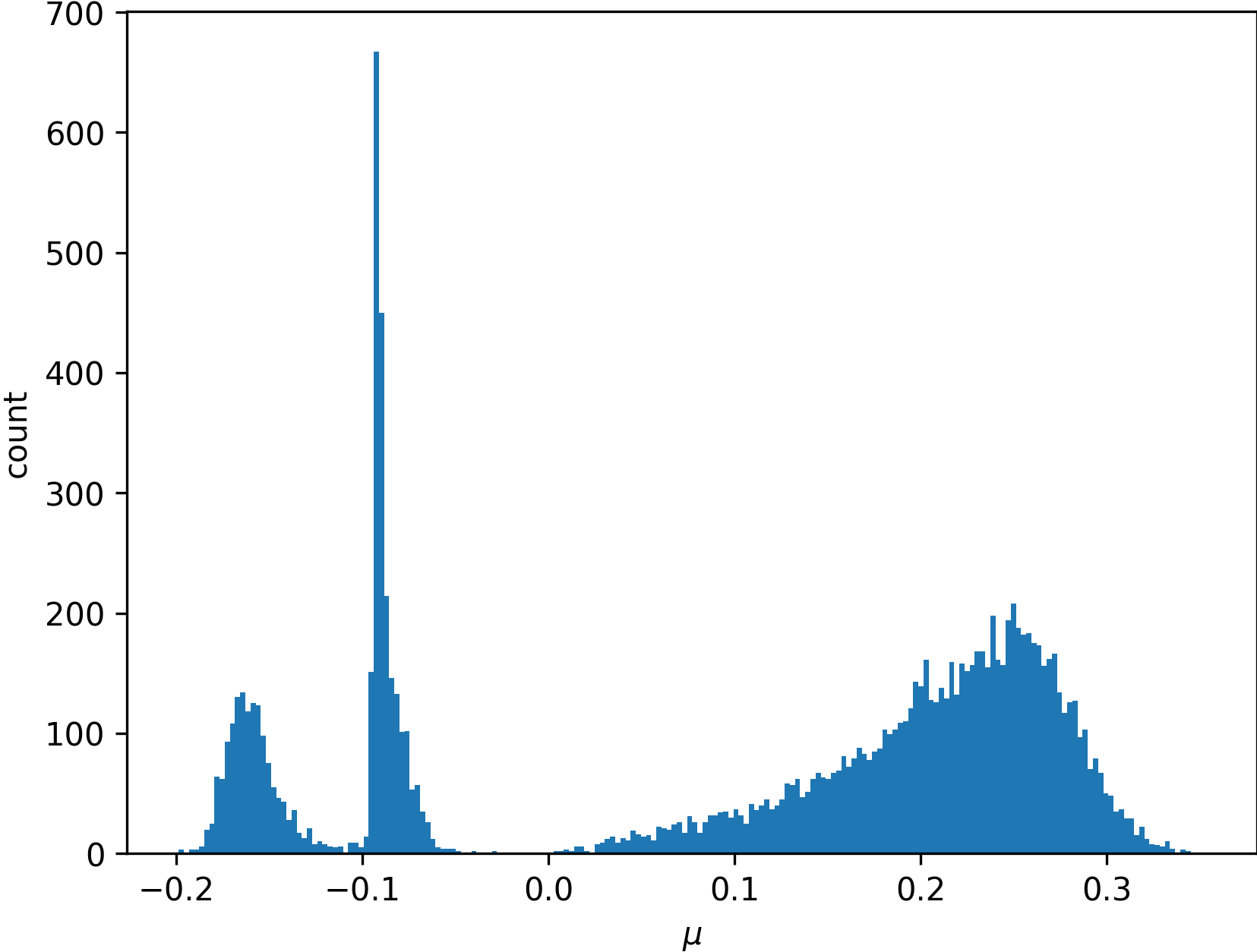} 
  \caption{Histogram of Convergence Measure, Granularity $10^{-11}$.}
  \label{fig:hist10m11}
\end{figure}

The statistical characteristics of the ($\mu_{att}$)-($\mu_{fc}$) landscape are nearly identical across different scales, as confirmed by the convergence-measure histograms. Figures~\ref{fig:hist10m5}, ~\ref{fig:hist10m8}, ~\ref{fig:hist10m10}, and ~\ref{fig:hist10m11} illustrate this property at resolutions of $10^{-5}$, $10^{-8}$, $10^{-10}$, and $10^{-11}$, respectively.

\begin{figure}[H]
  \centering
  \includegraphics[width=8cm]{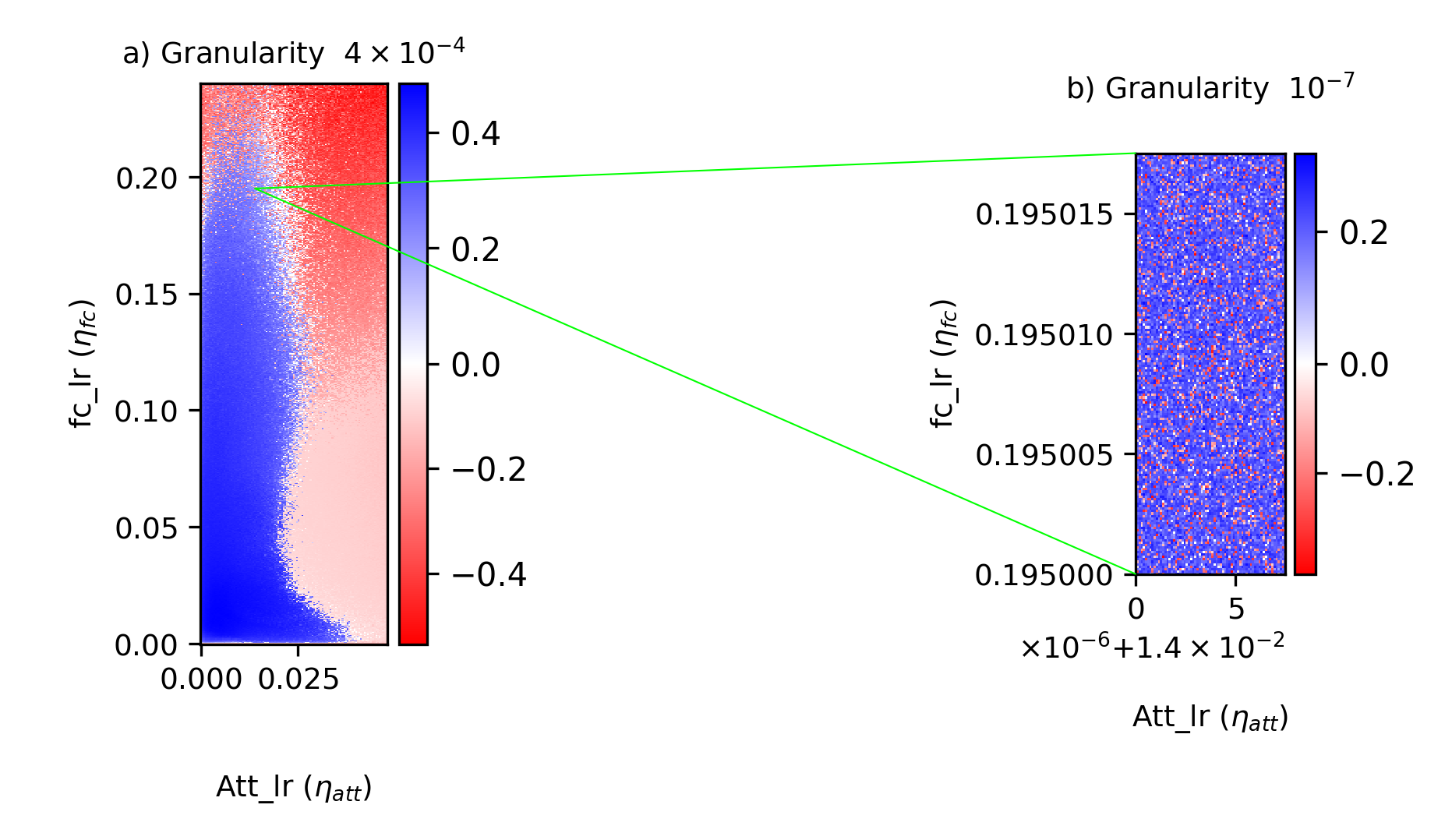} 
  \caption{Convergence Measure Heatmap – Boundaries at Granularity $10^{-7}$ Have a Dimension of $1.8118$.}
  \label{fig:zoomed_14}
\end{figure}

\begin{figure}[H]
  \centering
  \includegraphics[width=8cm]{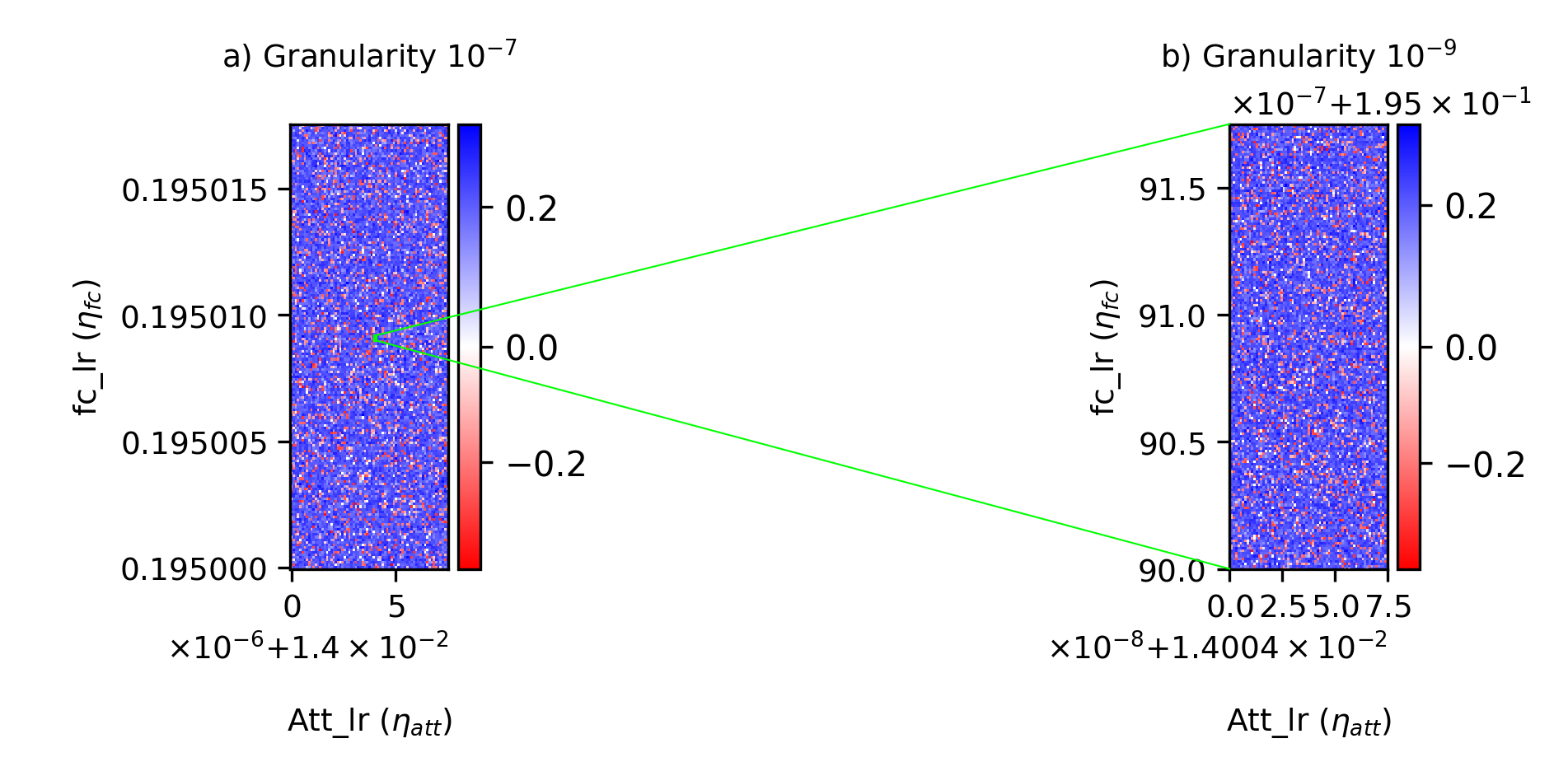} 
  \caption{Convergence Measure Heatmap – Boundaries at Granularity $10^{-9}$ Have a Dimension of $1.8218$.}
  \label{fig:zoomed_15}
\end{figure}

\begin{figure}[H]
  \centering
  \includegraphics[width=6cm]{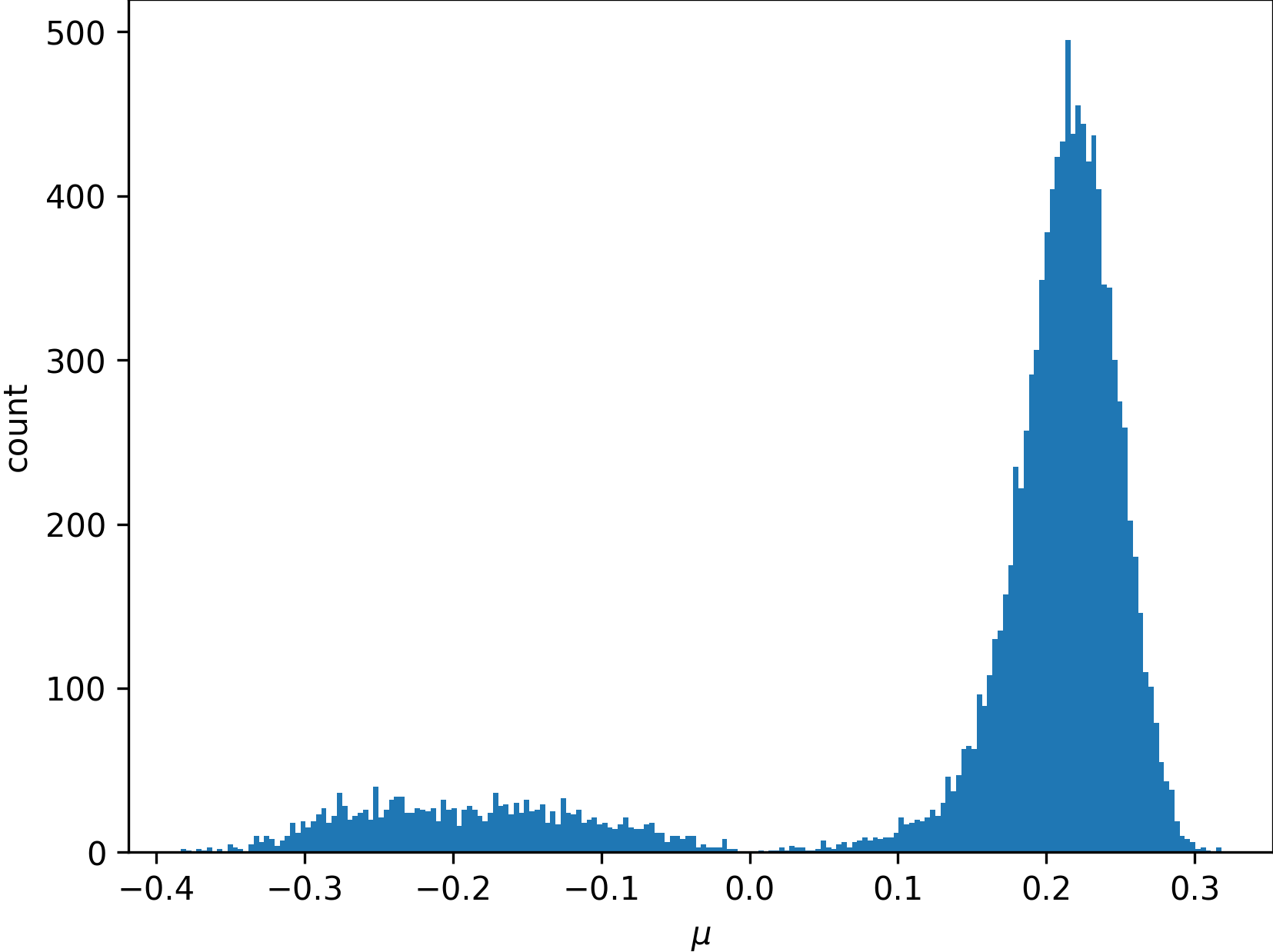} 
  \caption{Histogram of Convergence Measure, Granularity $10^{-7}$.}
  \label{fig:hist5}
\end{figure}

\begin{figure}[H]
  \centering
  \includegraphics[width=6cm]{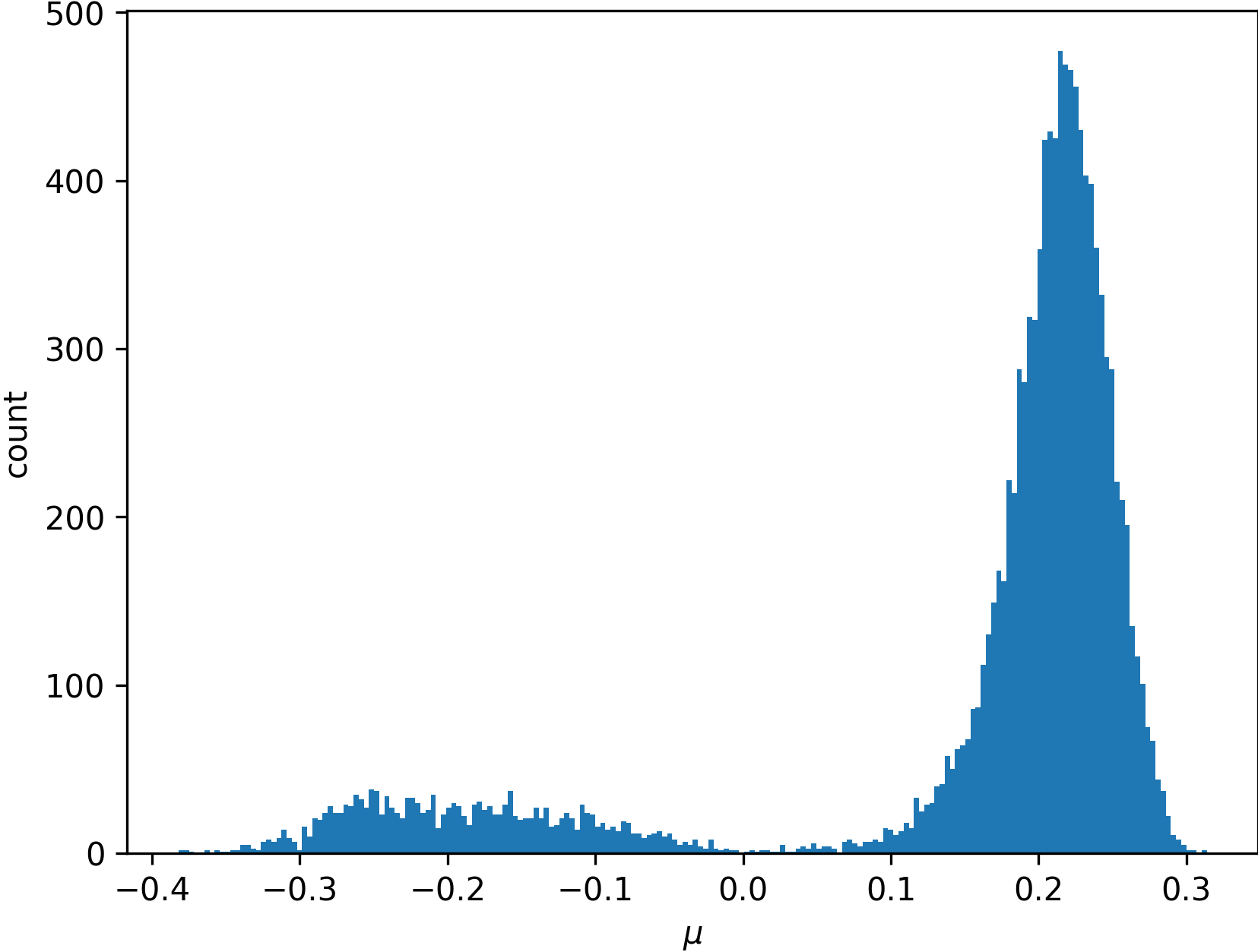} 
  \caption{Histogram of Convergence Measure, Granularity $10^{-9}$.}
  \label{fig:hist6}
\end{figure}

\begin{figure}[H]
  \centering
  \includegraphics[width=8cm]{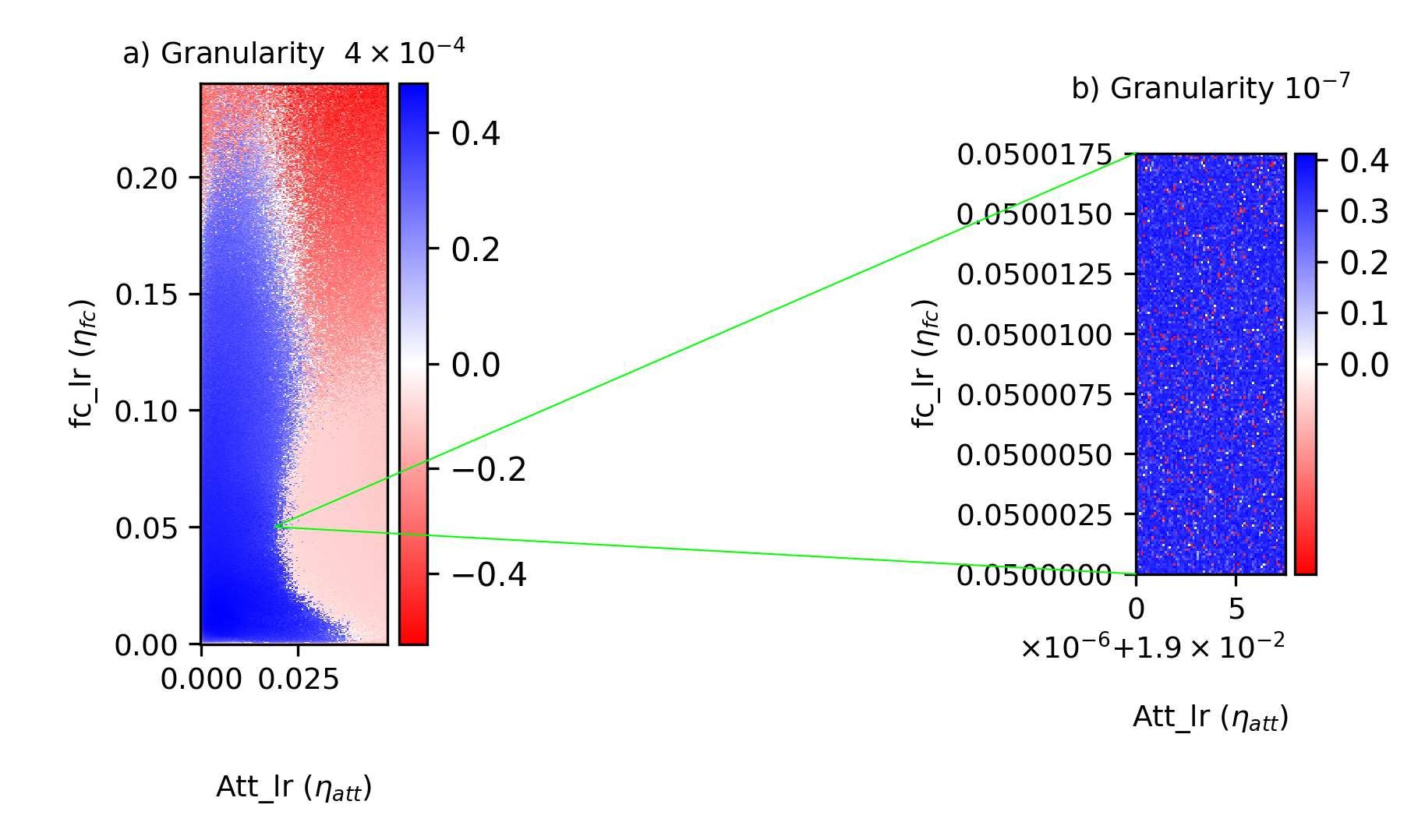} 
  \caption{Convergence Measure Heatmap – Boundaries at Granularity $10^{-7}$ Have a Dimension of $1.5810$.}
  \label{fig:zoomed_16}
\end{figure}

\begin{figure}[H]
  \centering
  \includegraphics[width=8cm]{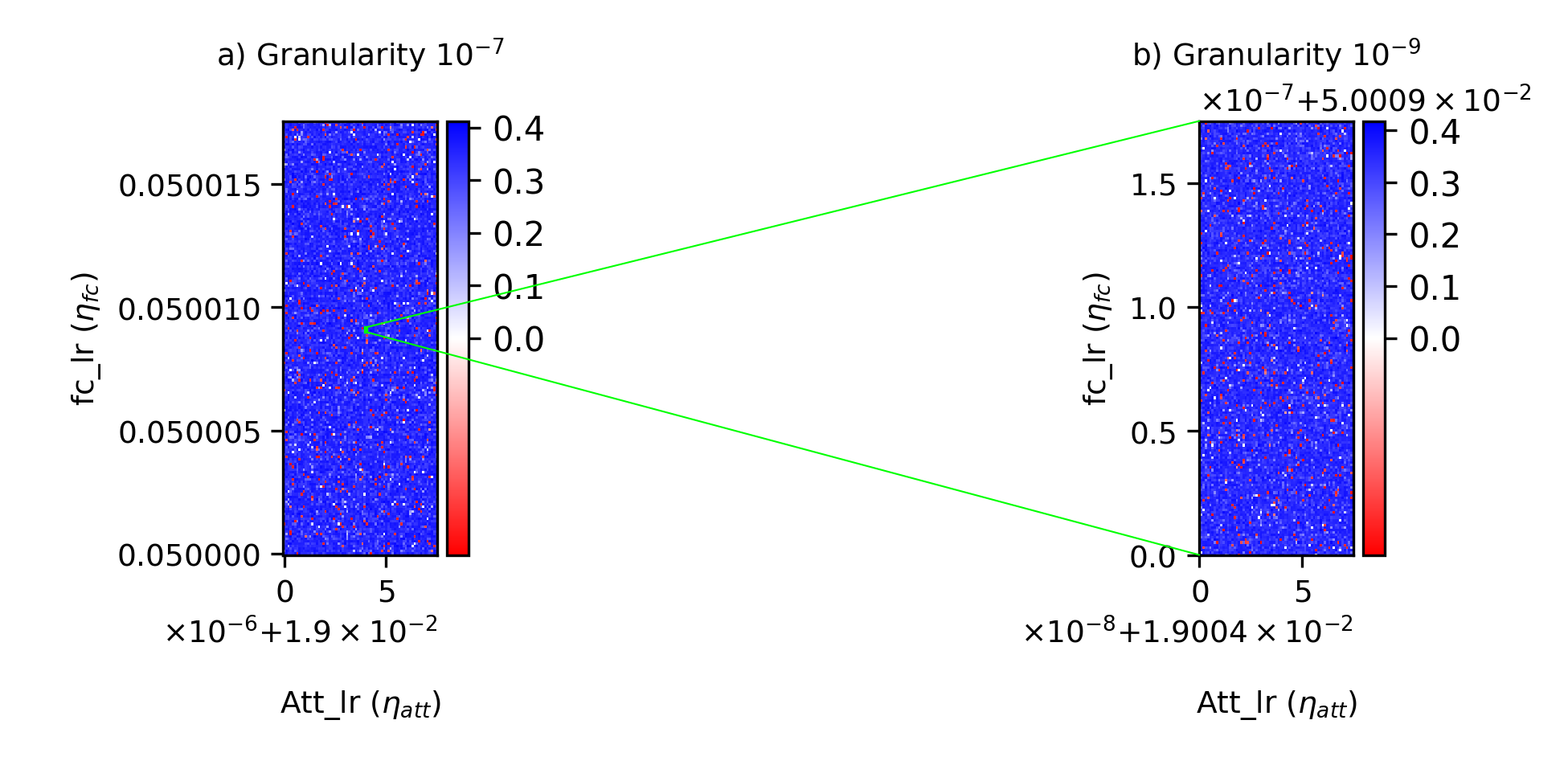} 
  \caption{Convergence Measure Heatmap – Boundaries at Granularity $10^{-9}$ Have a Dimension of $1.5413$.}
  \label{fig:zoomed_17}
\end{figure}

\begin{figure}[H]
  \centering
  \includegraphics[width=6cm]{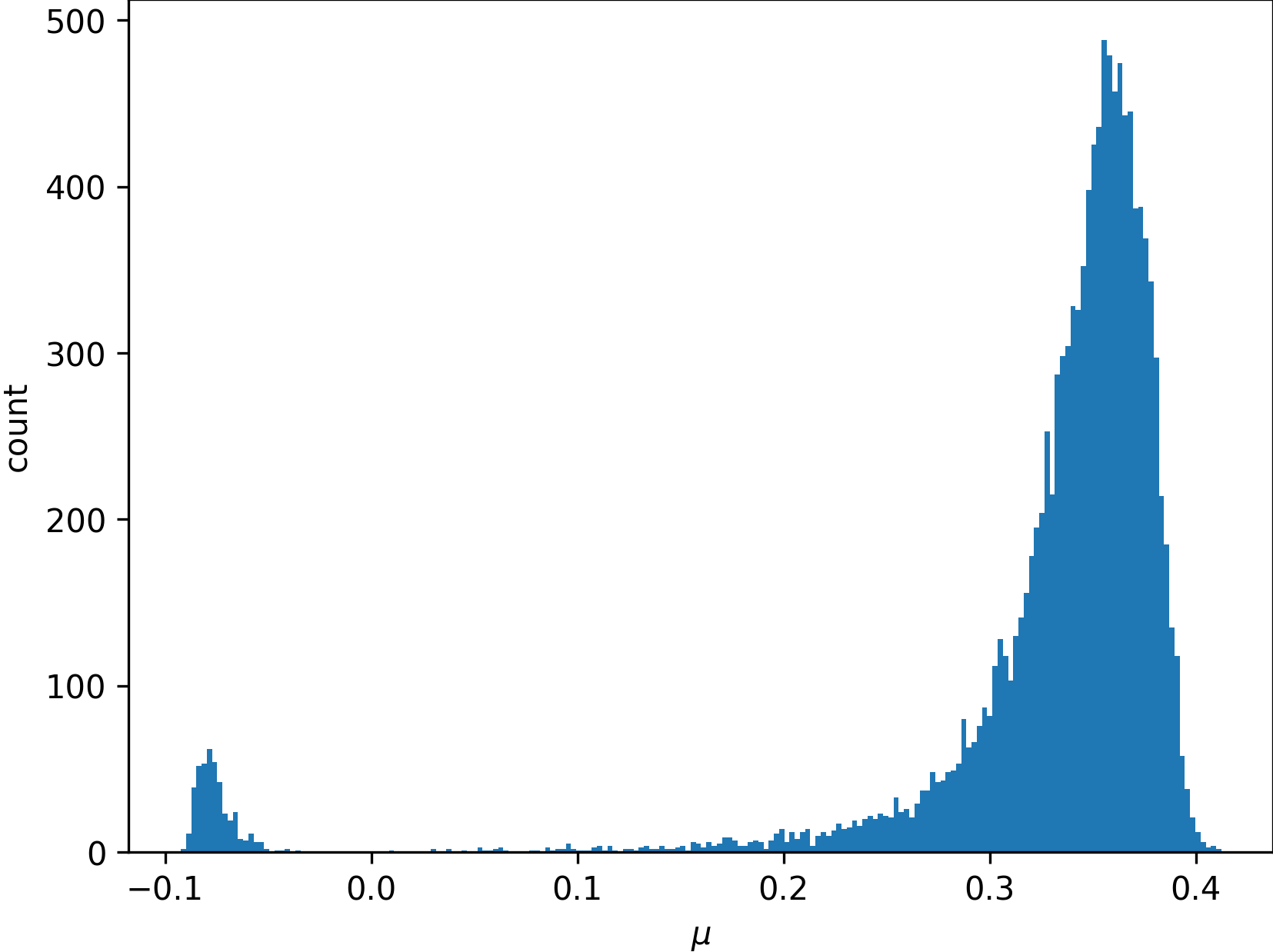} 
  \caption{Histogram of Convergence Measure, Granularity $10^{-7}$.}
  \label{fig:hist7}
\end{figure}

\begin{figure}[H]
  \centering
  \includegraphics[width=6cm]{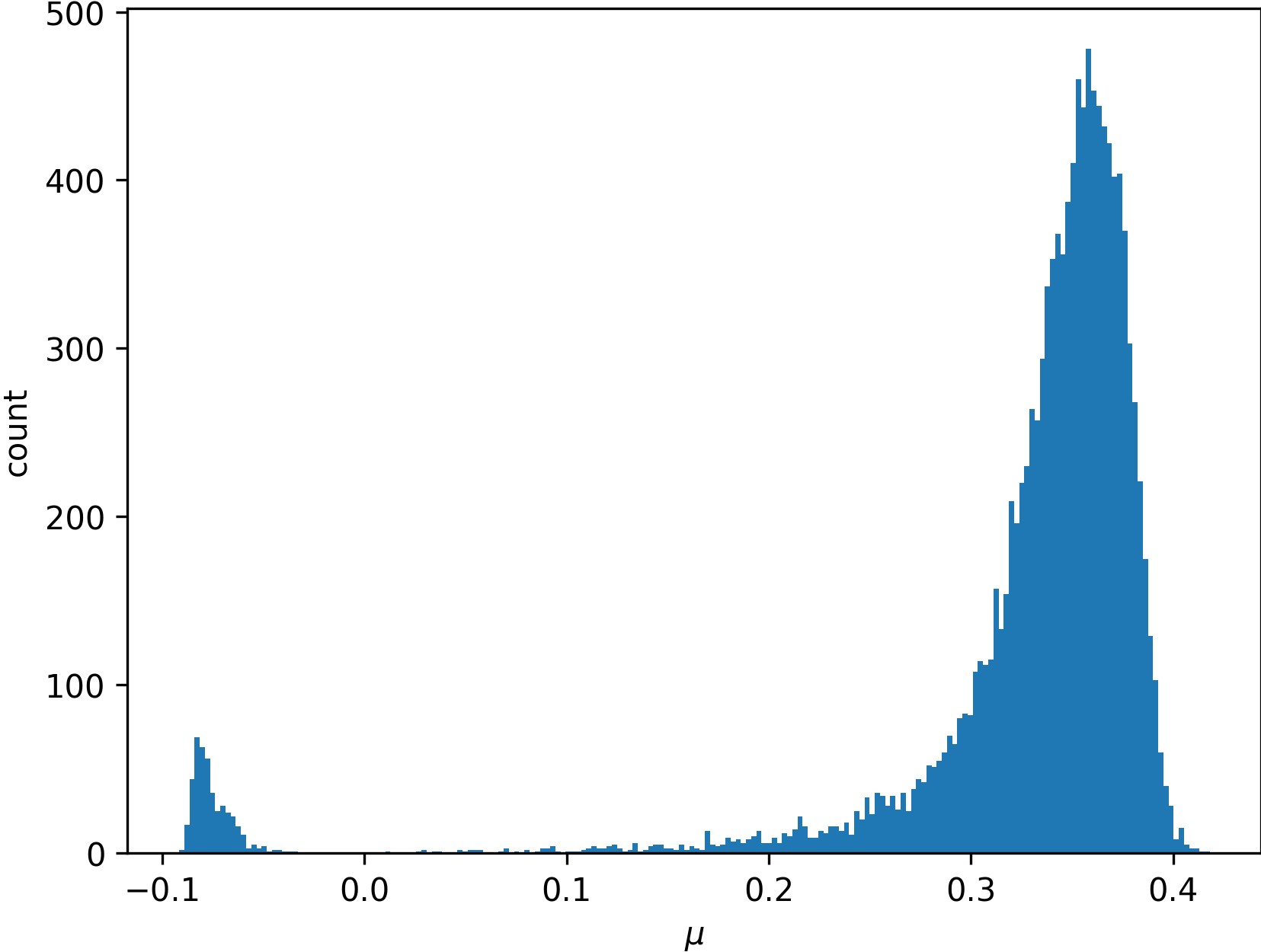} 
  \caption{Histogram of Convergence Measure, Granularity $10^{-9}$.}
  \label{fig:hist8}
\end{figure}

For completeness, I conducted the same experiments in two additional boundary regions. The findings again revealed similar statistical self-similarity, with non-integer fractal dimensions, at the upper portion of the region separating divergent and convergent behaviors (Figures~\ref{fig:zoomed_14}, ~\ref{fig:zoomed_15}, ~\ref{fig:hist5}, and \ref{fig:hist6}). Boundaries in this area have fractal dimensions of $1.8118$ and $1.8218$, respectively. Further inspection of the lower portion (Figures\ref{fig:zoomed_16}, ~\ref{fig:zoomed_17}, ~\ref{fig:hist7}, and ~\ref{fig:hist8}) also confirmed the presence of the same phenomenon, with dimensions $1.5810$
and $1.5413$.
Overall, these results suggest a highly chaotic boundary between convergence and divergence in transformer-based language models with statistical self similarity at different scales.

%%%%%%%%%%%%%%%%%%%%%
\pagebreak

%###############

\pagebreak
\pagebreak
%%%%%%%%%%%%%%%%%%%%%%%%%%%%%%%%%%%%%%%%%%%%%%%%%%%%%

\section{Conclusion}

This study investigated the nature of hyperparameter-space boundaries separating stable from divergent training regimes in a decoder-only transformer architecture. Drawing parallels to fractal-generating iterative processes observed in smaller networks, a more consistent convergence measure was introduced to visualize the training landscape over various learning rates for attention and fully connected layers. The results revealed repeating patterns across multiple scales, exhibiting chaotic features and self-similarity, nearly identical statistical characteristics, as evidenced by histograms of convergence measure and non-integer box-counting dimensions.

Owing to constrained training conditions, data, and computational resources, exploring more expansive hyperparameter spaces or larger models was not feasible. Nonetheless, these findings open promising avenues for further research. Expanding these experiments to more complex model architectures, larger datasets, and more diverse optimization algorithms could help determine whether such intricate and scale-invariant boundaries are intrinsic features of modern large-scale deep learning.

\clearpage

\bibliographystyle{unsrt}  
\bibliography{references}

\begin{thebibliography}{10}

\bibitem{sohl2024boundary}
Jascha Sohl-Dickstein.
\newblock The boundary of neural network trainability is fractal.
\newblock {\em arXiv preprint arXiv:2402.06184}, 2024.

\bibitem{barnsley2014fractals}
Michael~F Barnsley.
\newblock {\em Fractals everywhere}.
\newblock Academic press, 2014.

\bibitem{falconer2013fractal}
Kenneth Falconer.
\newblock {\em Fractal geometry: mathematical foundations and applications}.
\newblock John Wiley \& Sons, 2013.

\bibitem{kaye1994random}
Brian~H. Kaye.
\newblock {\em A Random Walk Through Fractal Dimensions}.
\newblock VCH Verlagsgesellschaft / VCH Publishers, 2nd edition, 1994.

\bibitem{peitgen2004chaos}
Heinz-Otto Peitgen, Hartmut J{\"u}rgens, Dietmar Saupe, and Mitchell~J
  Feigenbaum.
\newblock {\em Chaos and fractals: new frontiers of science}, volume 106.
\newblock Springer, 2004.

\bibitem{kingma2014adam}
Diederik~P. Kingma and Jimmy Ba.
\newblock Adam: A method for stochastic optimization.
\newblock {\em arXiv preprint arXiv:1412.6980}, 2014.

\bibitem{sohl-dickstein_fractal}
Jascha Sohl-Dickstein.
\newblock fractal.
\newblock \url{https://github.com/Sohl-Dickstein/fractal}, 2024.
\newblock Accessed: 2024-12-02.

\bibitem{vaswani2017attention}
Ashish Vaswani, Noam Shazeer, Niki Parmar, Jakob Uszkoreit, Llion Jones,
  Aidan~N. Gomez, Lukasz Kaiser, and Illia Polosukhin.
\newblock Attention is all you need.
\newblock {\em Advances in Neural Information Processing Systems}, 2017.

\bibitem{radford2018improving}
Alec Radford, Karthik Narasimhan, Tim Salimans, and Ilya Sutskever.
\newblock Improving language understanding by generative pre-training.
\newblock
  \url{https://cdn.openai.com/research-covers/language-unsupervised/language_understanding_paper.pdf},
  2018.
\newblock Accessed: 2024-04-27.

\bibitem{radford2019language}
Alec Radford, Jeffrey Wu, Rewon Child, David Luan, Dario Amodei, and Ilya
  Sutskever.
\newblock Language models are unsupervised multitask learners.
\newblock
  \url{https://cdn.openai.com/better-language-models/language_models_are_unsupervised_multitask_learners.pdf},
  2019.
\newblock Accessed: 2024-04-27.

\bibitem{brown2020language}
Tom~B. Brown, Benjamin Mann, Nick Ryder, Melanie Subbiah, Jared~D. Kaplan,
  Prafulla Dhariwal, Arvind Neelakantan, Pranav Shyam, Girish Sastry, Amanda
  Askell, Sandhini Agarwal, Ariel Herbert-Voss, Gretchen Krueger, Tom Henighan,
  Rewon Child, Aditya Ramesh, Daniel~M. Ziegler, Jeffrey Wu, Clemens Winter,
  Christopher Hesse, Mark Chen, Eric Sigler, Mateusz Litwin, Scott Gray,
  Benjamin Chess, Jack Clark, Christopher Berner, Sam McCandlish, Alec Radford,
  Ilya Sutskever, and Dario Amodei.
\newblock Language models are few-shot learners, 2020.

\bibitem{alexey2021image}
Alexey Dosovitskiy, Lucas Beyer, Alexander Kolesnikov, Dirk Weissenborn,
  Xiaohua Zhai, Thomas Unterthiner, Mostafa Dehghani, Matthias Minderer, Georg
  Heigold, Sylvain Gelly, Jakob Uszkoreit, and Neil Houlsby.
\newblock An image is worth 16x16 words: Transformers for image recognition at
  scale.
\newblock {\em arXiv preprint arXiv:2010.11929}, 2021.

\bibitem{dong2018speech}
Linhao Dong, Shuang Xu, and Bo~Xu.
\newblock Speech-transformer: a no-recurrence sequence-to-sequence model for
  speech recognition.
\newblock In {\em 2018 IEEE international conference on acoustics, speech and
  signal processing (ICASSP)}, pages 5884--5888. IEEE, 2018.

\bibitem{you2019large}
Yang You, Jing Li, Jonathan Hseu, Xiaodan Song, James Demmel, and Cho-Jui
  Hsieh.
\newblock Reducing bert pre-training time from 3 days to 76 minutes.
\newblock {\em arXiv preprint arXiv:1904.00962}, 12:2, 2019.

\bibitem{shakespeare_complete_works}
William Shakespeare.
\newblock {\em The Complete Works of William Shakespeare}.
\newblock Project Gutenberg, 1994.
\newblock Accessed: 2024-12-02.

\end{thebibliography}

\newpage

\section{Appendix}

\begin{figure}[H] % Use [H] to force the exact placement
  \centering
  \begin{minipage}{0.45\textwidth}
    \centering
    \includegraphics[width=\linewidth]{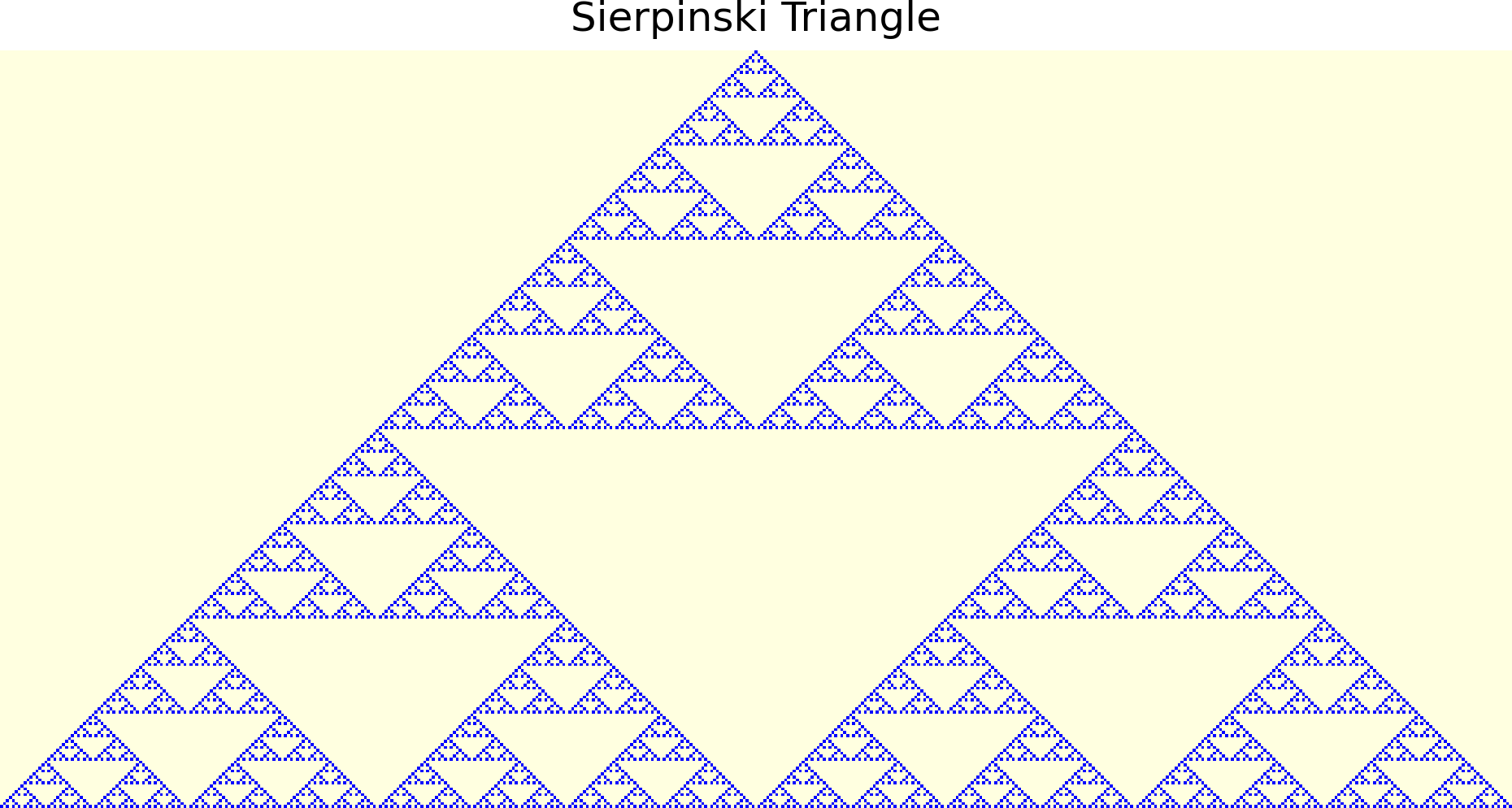}
    \caption{Sierpinski Fractal with Theoretical Dimension of 1.585.}
    \label{fig:Sierpinski}
  \end{minipage}
  \hfill
  \begin{minipage}{0.45\textwidth}
    \centering
    \includegraphics[width=\linewidth]{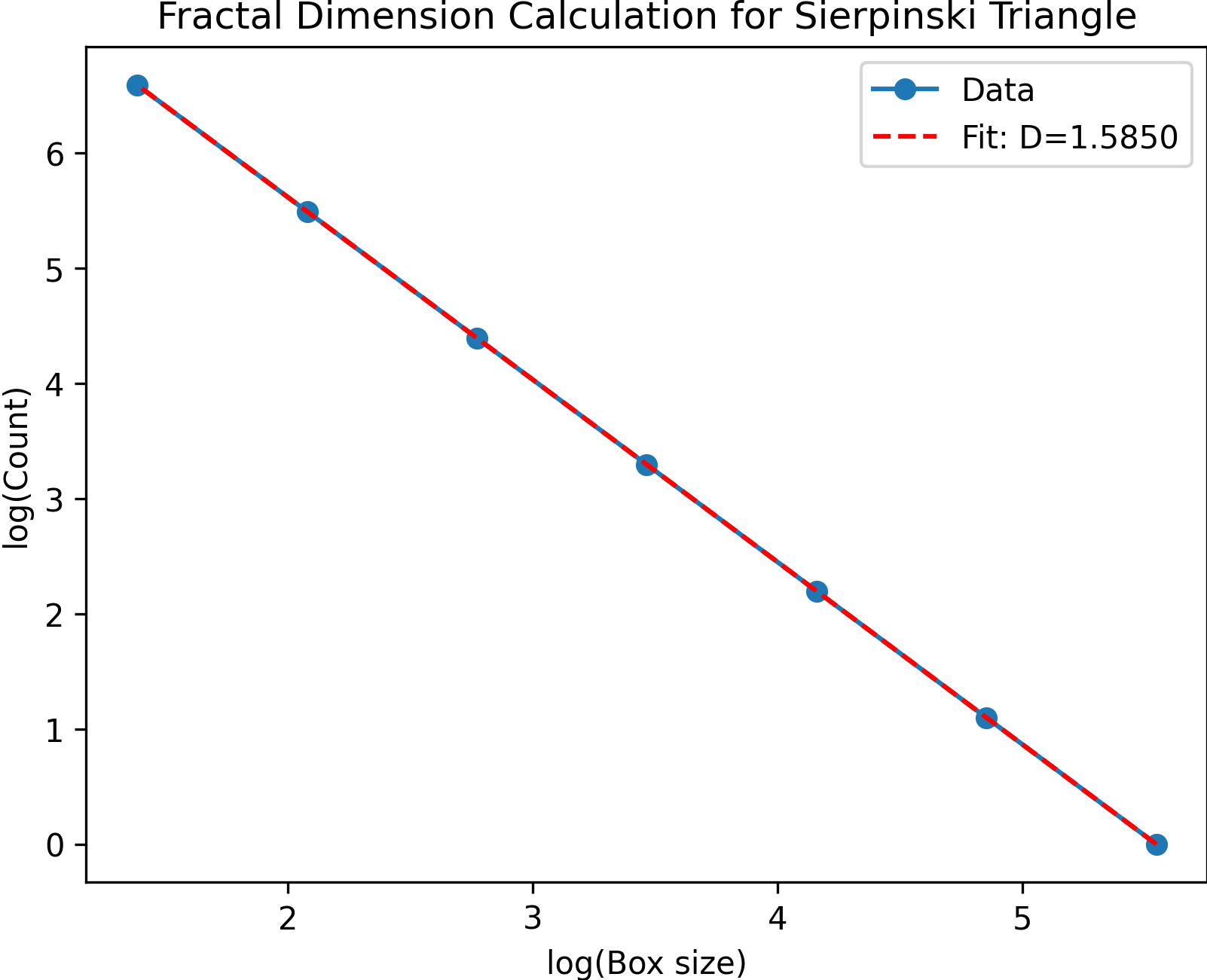}
    \caption{Fractal Dimension Estimated as 1.5850, Using Box-Counting Technique.}
    \label{fig:Box-Counting}
  \end{minipage}
\end{figure}

\begin{figure}[H] % Use [H] to force the exact placement
  \centering
  \begin{minipage}{0.45\textwidth}
    \centering
    \includegraphics[width=\linewidth]{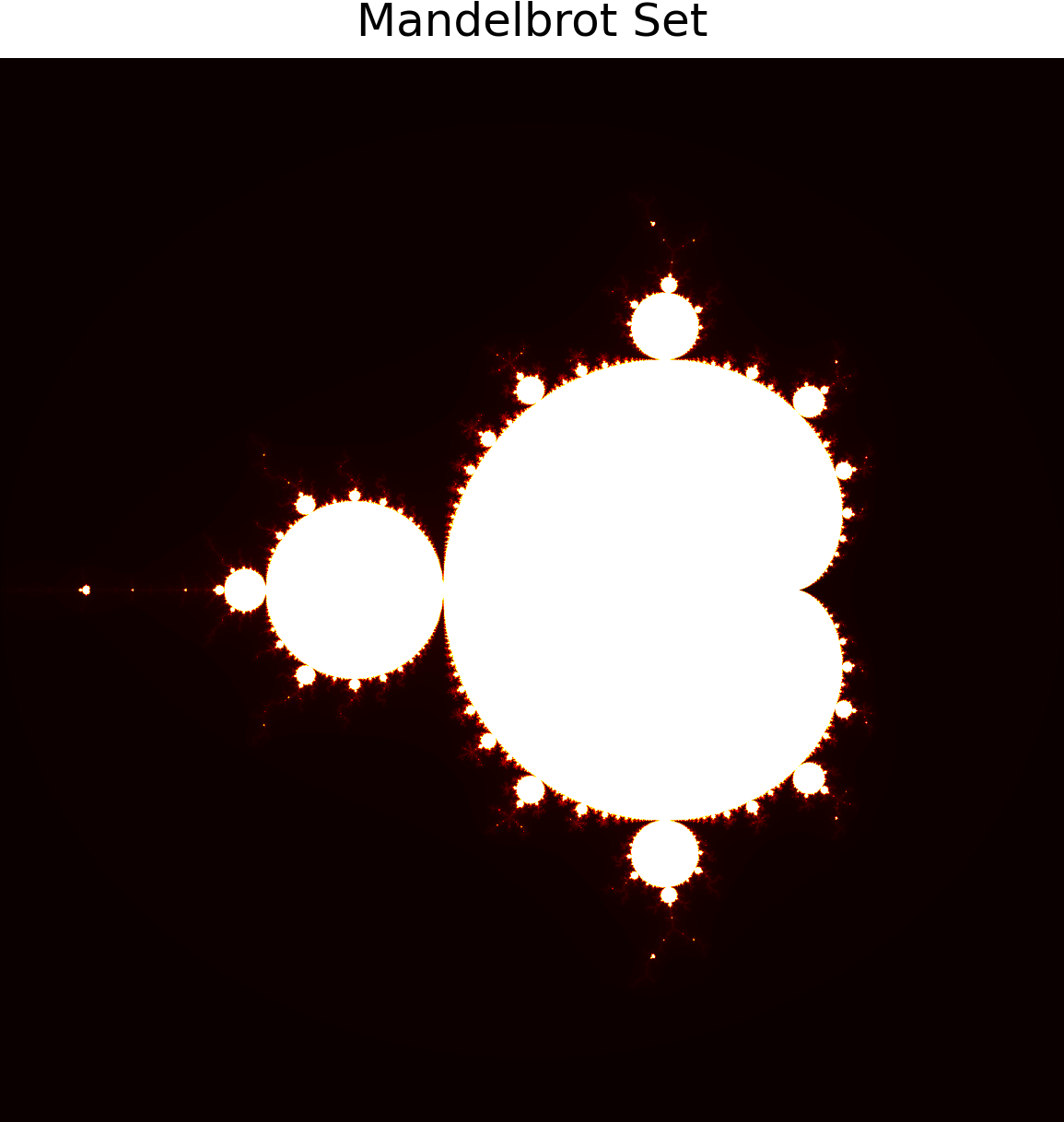}
    \caption{Mandelbrot Set.}
    \label{fig:Mandelbrot}
  \end{minipage}
  \hfill
  \begin{minipage}{0.45\textwidth}
    \centering
    \includegraphics[width=\linewidth]{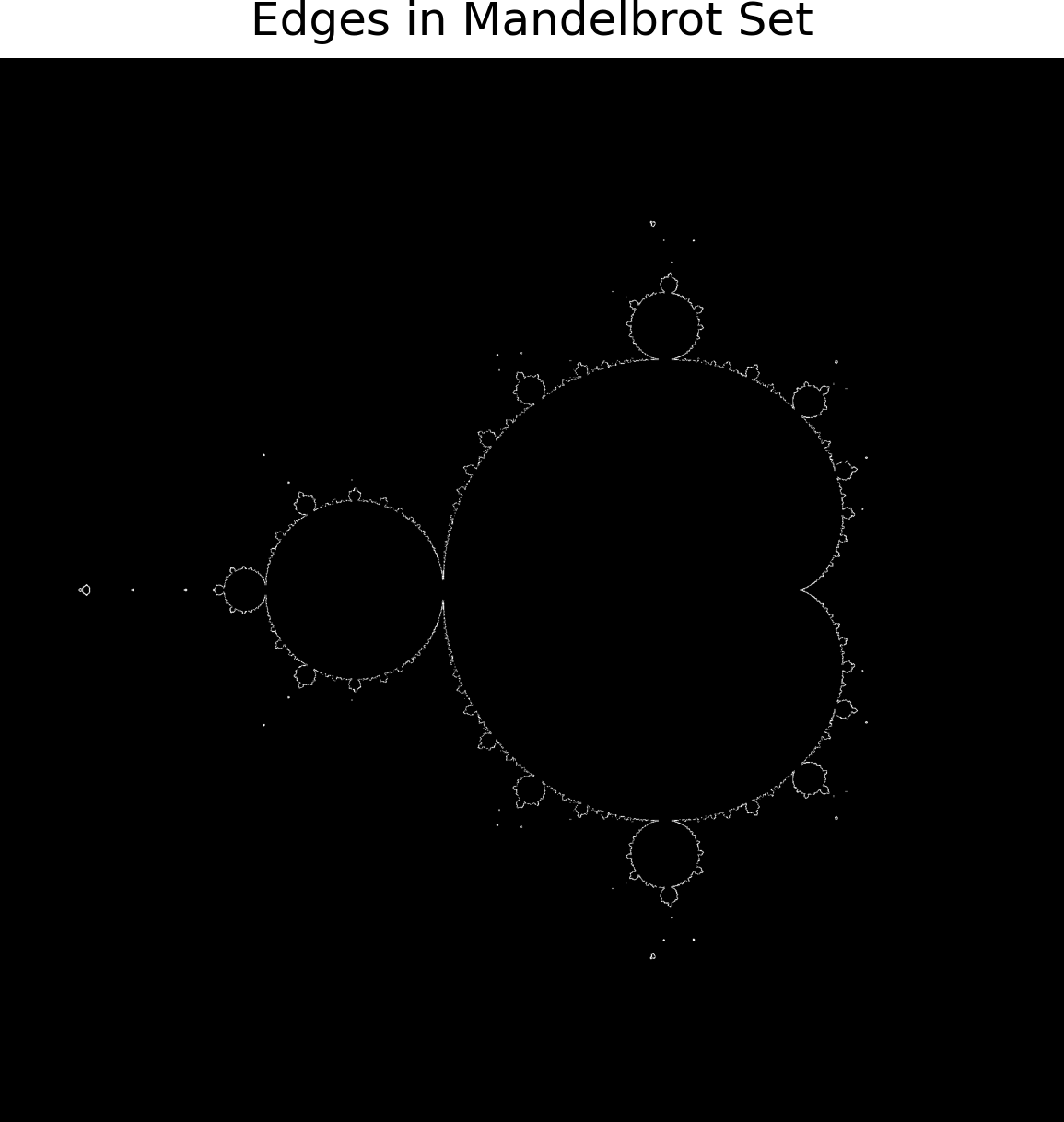}
    \caption{Edges Detected by Sobel.}
    \label{fig:Sobel}
  \end{minipage}
\end{figure}

I use box-counting method to estimate fractal dimensions and Sobel operator to detect edges. For verification porpuses, I applied the box-counting technique to Sierpinski fractal with dimension 1.585 and tested edge detection by identifying the edges of the Mandelbrot set. Both results are presented in figures ~\ref{fig:Sierpinski}, ~\ref{fig:Box-Counting}, ~\ref{fig:Mandelbrot} and ~\ref{fig:Sobel}. 

%###############

By applying those techniques to a black and white version of the hyperprameter landscape, I was able to calculate the fractal dimension for different levels of magnification. 
\footnote{Refer to the code written in Python using JAX and Flax libraries at \url{https://github.com/tbahman/MAPPING_THE_EDGE_OF_CHAOS}.}

\end{document}